\documentclass[letterpaper, 10 pt, journal, twocolumn]{IEEEtran}
\usepackage[utf8]{inputenc}
\usepackage{bm}
\usepackage{amsmath}
\usepackage{amsthm}
\usepackage{amssymb}
\usepackage{graphicx}
\usepackage{stfloats}
\newtheorem{assumption}{Assumption}

\newtheorem{remark}{Remark}

\newtheorem{condition}{Condition}

\newcommand{\mbf}[1]{\mathbf{#1}}

\newcommand{\mbfh}[1]{\mathbf{\hat{#1}}} 
\newcommand{\mbfd}[1]{\dot{\mathbf{#1}}}

\newcommand{\mcl}[1]{\mathcal{#1}}

\newcommand{\Real}{\mathbb{R}} 
\newcommand{\RealPos}{\mathbb{R}_{+}} 

\newcommand{\xv}{\mbf{x}}

\newcommand{\zvh}{\mbfh{z}}
\newcommand{\qvd}{\mbfd{q}}

\newcommand{\pv}{\mbf{p}}

\newcommand{\fv}{\mbf{f}}

\newcommand{\wv}{\mbf{w}}

\newcommand{\dv}{\mbf{d}}

\newcommand{\Lv}{\mbf{L}}

\newcommand{\lambdav}{\bm{\lambda}}
\renewcommand{\baselinestretch}{.99}

\IEEEoverridecommandlockouts

\title{Personalized On-line Adaptation of Kinematic Synergies for Human-Prosthesis Interfaces}
\author{Ricardo Garcia-Rosas, Ying Tan, Denny Oetomo, Chris Manzie, and Peter Choong \thanks{This project is funded by the Valma Angliss Trust.} \thanks{R. Garcia-Rosas, Y. Tan, D. Oetomo, and C. Manzie are with the School of Electrical, Mechanical and Infrastructure Engineering, and P. Choong with the Department of Surgery, The University of Melbourne, VIC 3010, Australia. {\tt ricardog@student.unimelb.edu.au; \{yingt,doetomo,manziec,pchoong\}@unimelb.edu.au}.}}
\begin{document}

%First page footnote
\maketitle

\begin{abstract}
Synergies have been adopted in prosthetic limb applications to reduce complexity of design, but typically involve a single synergy setting for a population and ignore individual preference or adaptation capacity. However, personalization of the synergy setting is necessary for the effective operation of the prosthetic device. Two major challenges hinder the personalization of synergies in human-prosthesis interfaces. The first is related to the process of human motor adaptation and the second to the variation in motor learning dynamics of individuals. In this paper, a systematic personalization of kinematic synergies for human-prosthesis interfaces using on-line measurements from each individual is proposed. 
The task of reaching using the upper-limb is described by an objective function and the interface is parameterized by a kinematic synergy. Consequently, personalizing the interface for a given individual can be formulated as finding an optimal personalized parameter. A structure to model the observed motor behavior that allows for the personalized traits of motor preference and motor learning is proposed, and subsequently used in an on-line optimization scheme to identify the synergies for an individual. The knowledge of the common features contained in the model enables on-line adaptation of the human-prosthesis interface to happen concurrently to human motor adaptation without the need to re-tune the personalization algorithm for each individual.
Human-in-the-loop experimental results with able-bodied subjects, performed in a virtual reality environment to emulate amputation and prosthesis use, show that the proposed personalization algorithm was effective in obtaining optimal synergies with a fast uniform convergence speed across a group of individuals. 
\end{abstract}


% Introduction
%
%
%
\section{Introduction}
In recent years, synergistic Human-Prosthesis Interfaces (referred to as HPIs henceforth) have been actively studied due to their potential to realize the coordinated motion between the residual-limb and prosthesis required by most activities of the daily living \cite{Gottlieb1996,Cordella2016}. Reaching for an item on a shelf requires the shoulder, elbow and wrist joints to move in a coordinated fashion as opposed to a sequential joint by joint motion seen in traditional prostheses.  In synergistic HPIs, a set of parameters are used to determine the prosthesis' motion in response to the motion of the user's residual limb \cite{Alshammary2018, Legrand2018}. These parameters are referred to as synergy herein.

Most of the current kinematic synergy-based HPI strategies use artificial neural network and regression based approaches, where the data gathered from able-bodied subjects is used to train the algorithms \cite{Kaliki2013, Blana2016, Merad2016a, Akhtar2017}. This yields the interface setting that best fits the population from which the training data-set was created. However, given individual variations in motor preferences, capabilities and the effects of motor learning, as well as the more significant changes due to amputation, these interfaces needed to be systematically personalized to be effective for amputees \cite{Merad2018}. In \cite{Legrand2018}, different techniques of linear regression were performed in a bid to obtain a common synergy for a group of able-bodied people. The identified synergy setting was then used on individual emulated amputees. This study concluded that the choice of regression technique did not significantly improve the performance of the synergistic prosthetic interface. 
The results in \cite{Legrand2018}, however, demonstrated that while general common motor features can be observed (a general trend in limb trajectory), there is also a significant amount of inter-subject motor variability.

In order to accommodate human-to-human physiological and motor behavior variations, and the numerous variations in the nature of amputation \cite{Biddiss2007a}, personalization of HPIs is a necessary and critical step in realizing the effective operation of a prosthesis \cite{Fougner2012}. This need for personalization has been explicitly identified for synergistic HPIs \cite{Merad2018}. It is because of these variations in motor behavior that there exists a personalized setting that produces the best performance, and that this personalized setting varies among individuals \cite{Garcia-Rosas2018EMBC}.

Two major challenges hinder the efforts of realizing personalized HPIs. 
%% Problem 1
The first challenge is related to the process of human motor adaptation. An individual needs to learn how to use the prosthesis for each given synergy setting. The best possible performance of the individual for a given setting is not known until the motor adaptation process has reached its steady-state, which may take many iterations of the task \cite{Garcia-Rosas2018EMBC}. Therefore, a significant amount of time will be required to evaluate all possible settings in an exhaustive process, which renders it impractical. 
%% Problem 2
The second challenge is related to the motor learning dynamics of individuals, which differ from person to person. Each individual learns at a different rate, which is initially unknown.
Determining that an individual has indeed reached the steady-state in motor learning, thus capturing the best performance for a given synergy setting, is challenging.  For example, it is insufficient to generalize that steady-state in motor learning is reached after a set number of iterations for every individual \cite{Garcia-Rosas2018EMBC}.

This paper seeks to address the two challenges presented above to allow the systematic personalization of synergistic HPIs. A model-guided data-driven optimization approach is proposed, providing a systematic method to explore and continuously adjust the synergy setting towards a personal optimum, avoiding the need for exhaustive testing or calibration. The suggested method enables the identification of the best synergy setting to take place while the prosthetic device is being used to perform a task, a process referred to as ``on-line personalization'' herein. 
An example of the on-line personalization of synergistic HPIs through a data-driven optimization procedure was presented in the authors' previous work \cite{Garcia-Rosas2018ACC}, where the individual's performance is given by a pre-defined objective function that is evaluated from sensor measurements in the prosthetic device. However, the method presented in \cite{Garcia-Rosas2018ACC} suffers from the previously discussed challenges as it requires the user to undergo significant previous training with the synergistic prosthesis in order to treat the personalization of synergistic HPIs as a static optimization problem, i.e. the algorithm did not account for the transient dynamics of human motor learning and the algorithm learning rate had to be manually adjusted for each individual. Similar attempts have been performed in lower limb prostheses \cite{Kumar2017}.

The method proposed in this paper enables the on-line personalization of synergistic HPIs to operate concurrently to the individual learning to use the prosthetic device. This bypasses the need for the algorithm to wait for human motor learning to reach steady-state at each synergy setting before evaluating the performance in a new setting. This also avoids the need for the algorithm to be tuned for each individual. This is achieved by realizing that the human motor behavior involved can be captured in a model composed of a non-linear mapping between individual performance and the given synergy setting, and a linear time invariant system that describes the transient dynamics of human motor learning. This model then informs the design of a class of on-line optimization algorithm that takes advantage of the structure contained in the model independently of its parameters.

The paper is presented as follows. First, a grey-box model is proposed that captures the observed common human motor behavior in its structure and individual variation in its parameters. Then, an on-line optimization approach to the personalization of HPIs, realized through an extremum seeking-based algorithm \cite{Moase2011a, Moase2009a}, is presented; which takes advantage of the knowledge of human motor behavior that is captured by the grey-box model. Lastly, the experimental methodology and results for an experiment performed in virtual reality with able-bodied subjects are presented. The experiment demonstrates the proposed method can consistently adjust the prosthetic synergy setting to allow individuals to achieve their highest performance score as the prosthesis is continuously used. The proposed strategy is shown to be effective in personalizing a synergistic HPI to individuals learning to use the prosthetic device.

%
% Preliminaries
%
\section{Preliminaries}
%
% Notation
%
The following notation is used throughout the paper. The set of real numbers is denoted as $\Real$, the set of positive real numbers is denoted as $\RealPos$. The term ``synergy'' is used to describe the parametrization of the human-prosthesis interface. For a given static mapping $y_x = h(x):\Real\rightarrow \Real$, its gradient is expressed as $y^{'}_{x} = \partial h(x)/\partial x$, and curvature (Hessian) as $y^{''}_{x} = \partial^{2} h(x)/\partial x^{2}$. The subscript $x_{i}$ is used to indicate the iteration domain, such that $x_{i}$ indicates the $i$th iteration of $x$, where an iteration represents a repetition of the task. The ``bar'' notation on a variable, e.g. $\bar{\lambdav}$, indicates that the variable is individual dependent. The measured performance of a given task is characterized by the performance $\hat{J}_i\in\RealPos$. The synergy parameters are given by $\theta_i\in\Theta$, where $\Theta$ is a compact set in $\Real$ representing the parameter set of interest. 

A scalar linear relationship between the velocity of the shoulder and elbow flexion for the task of reaching forward is selected in this manuscript as the specific example of kinematic synergy to exemplify the application of the method presented herein. However, the proposed method is generalizable to other parametrizations of synergies of higher complexity and dimensionality. This scalar synergy, even though it is simple, has been found to be able to represent the most significant relationship between shoulder and elbow motion for forward reaching tasks \cite{Bockemuhl2010}, and has been successfully applied to prosthetic elbow studies \cite{Alshammary2018}. In this paper, the elbow to shoulder flexion kinematic synergy is described by

\vspace{-2mm}
\begin{equation}
\label{eq:linearKinematicSynergy}
    \dot{q}_{p} = \theta\dot{q}_{r},
    %\vspace{-2mm}
\end{equation}
where the subscripts $r$ and $p$ represent residual and prosthetic limb respectively. $\dot{q}_{r}$ is the individual's shoulder flexion angular velocity. $\dot{q}_{p}$ is the angular velocity of the prosthetic elbow.

Furthermore, the measured performance for a given reaching task in this paper is defined by the following objective function:
\begin{equation}
\label{eq:experimentalCost}
    \hat{J}_{i}(\theta) = \frac{0.25\|\tilde{\pv}_{fmax}\|^{2}}{\max{(0.25,\|\tilde{\pv}_{f,i}\|^{2}})} + \frac{16.67 t_{fmax}}{\max{(0.5,t_{f,i})}}.
\end{equation}
Here, $\tilde{\pv}_{f,i}$ is the hand end position error from target, and $\tilde{\pv}_{fmax} = 10 \textup{cm}$; $t_{f,i}$ is the reach time; and $t_{fmax} = 3 \textup{s}$. Note that both $\tilde{\pv}_{f,i}$ and $t_{f,i}$ are influenced by $\theta$ and $\qvd_r$, where $\qvd_r$ is the angular velocity of the residual limb. The values used in the objective function normalize its components to be in a $[0-100]$ range.
%
% Model of human motor preference and adaptation
%
\section{Human motor preference and adaptation model}
\label{sec:model}

In this section, the common features observed in motor behavior for individuals using a HPI based on kinematic synergies are discussed and their formalization within a grey-box model is proposed. Knowledge of the structure of human motor behavior captured by the grey-box model informs the design of the proposed personalization algorithm in Section \ref{sec:algorithm}, as well as its applicability to HPIs. Previously obtained experimental results showing these features \cite{Garcia-Rosas2018EMBC} are presented first, followed by the grey-box model proposed to characterize them. Lastly the experimental validation of the model is performed.

%\vspace{-1mm}
\subsection{Common features in human motor behavior}
It has been demonstrated that by characterizing an upper-limb reaching task, while using a synergistic prosthesis, by an objective function, equation (\ref{eq:experimentalCost}), and by parameterizing the HPI, equation (\ref{eq:linearKinematicSynergy}), two common motor features arise \cite{Garcia-Rosas2018EMBC}. The first feature will be referred to as ``motor preference'', which relates to the observed preference in synergy value that leads to the best performance of the task for a given individual. The second feature will be referred to as ``motor adaptation'' which represents an individual's improvement in performance over iterations of the task that eventually reaches a steady-state.

\subsubsection{Motor preference}
There exists a non-linear steady-state input-output relationship between synergies and a characterization of the task as an objective function (performance). This means that for a given individual and each synergy value, there will be specific steady-state performance (after learning) that the individual can achieve. This relationship is dependent on the formulation of the objective function, which can be designed such that the resultant synergy-performance map has an extremum \cite{Garcia-Rosas2018EMBC}.

\begin{figure}[ht]
    \centering
    \vspace{-2mm}
    \includegraphics[trim={0.0in 0.0in 0.0in 0.0in}, width=3.4in, clip]{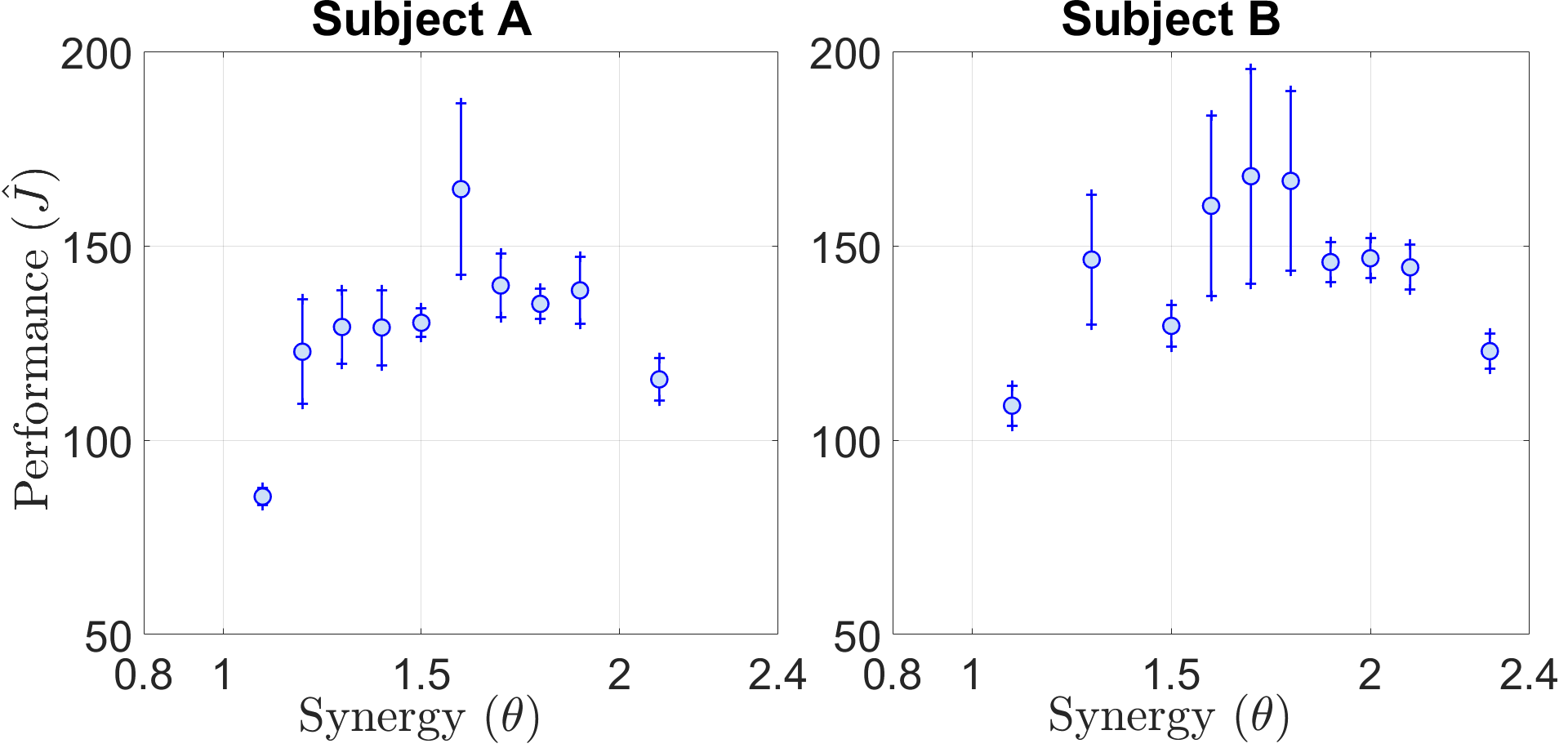}
    \vspace{-3mm}
    \caption{Experimentally obtained synergy-performance map for two representative subjects. The blue circles represent mean performance while the whiskers the standard deviation. Task performance was characterized by a quadratic-like objective function, while synergy represents a kinematic-synergy parametrization of the human-prosthesis interface.}
    \label{fig:sampleMaps}
    \vspace{-2mm}
\end{figure}

An example of the relationship between the synergy ($\theta$) and performance ($\hat{J}$) is shown in Figure \ref{fig:sampleMaps}. An individual's average performance over multiple iterations of the task for a range of synergy values conforms to an Input-to-Steady-State-Output (I-SSO) map whose extremum and shape differ across individuals. This individual variation in the best performing parameter suggests the existence of preference in the configuration of the prosthetic interface. The shape of this average map for individuals has been identified as the common feature while the parameters that define the map and its extremum as individual variations. In this work, the average synergy-performance map feature will be referred to as ``motor preference'', which represents an individual's preference in synergy parameter that leads to the best performance of the task. This implies that through the personalization of the synergy it is possible to improve the performance the prosthesis user can achieve when compared to the use of the population average synergy value.

\subsubsection{Motor adaptation} 
\begin{figure}[ht]
    \centering
    \vspace{-2mm}
    \includegraphics[trim={0.0in 0.0in 0.0in 0.0in}, width=3.4in, clip]{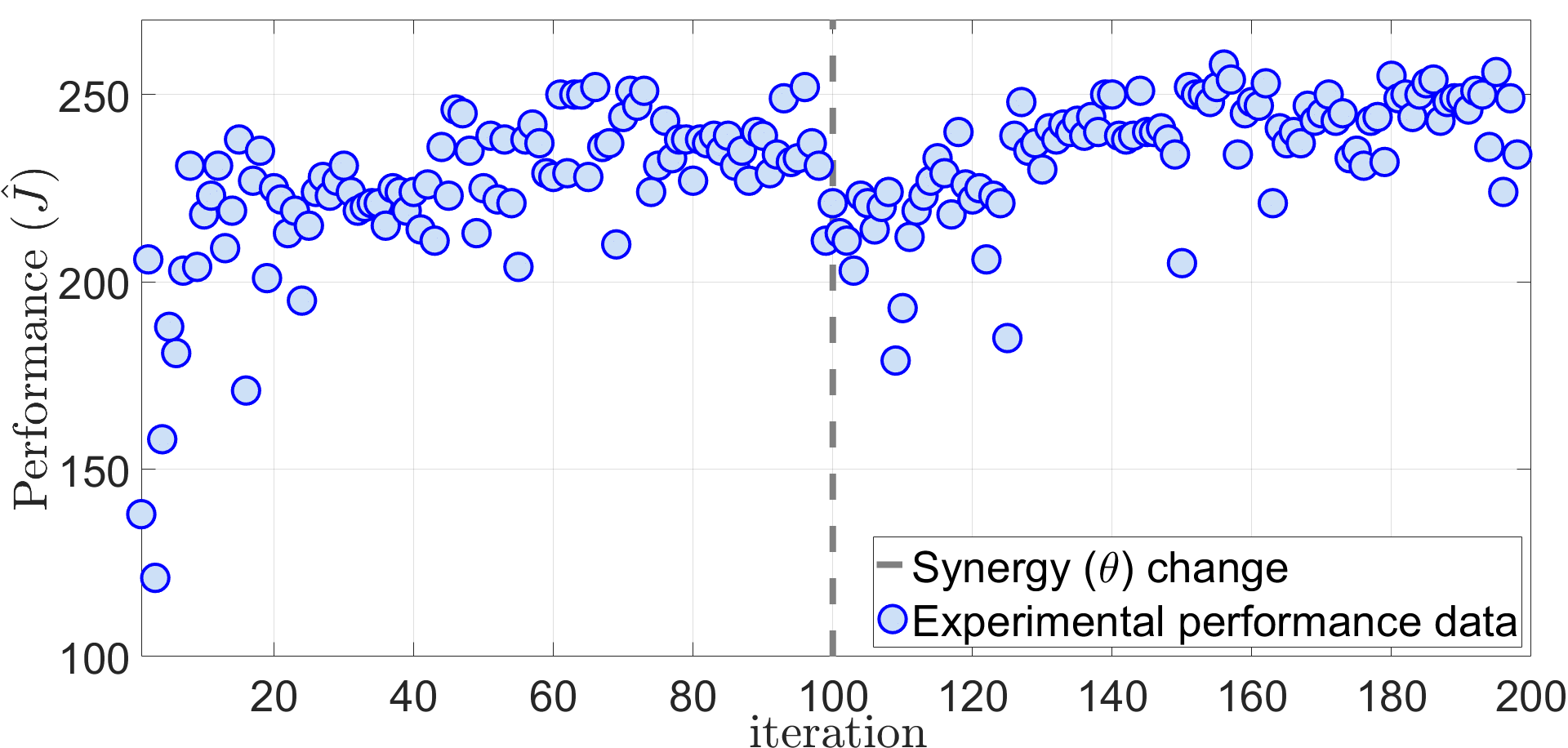}
    \vspace{-3mm}
    \caption{Experimentally obtained performance across iterations data for two synergy values for a representative subject. The dashed line represents the change in the value of the kinematic-synergy parametrization (synergy) of the human-prosthesis interface. Performance was characterized by a quadratic-like objective function.}
    \label{fig:sampleLTI}
    \vspace{-2mm}
\end{figure}

A transient behavior was observed in performance as the synergy setting was changed from one value to the next, where the dynamics are individual dependent. As expected from human behavior, the performance across iterations of the task is affected by the variability inherent in human motor control \cite{Riley2002}. An example of this transient behavior is presented in Figure \ref{fig:sampleLTI}, where the transient behavior can be seen before the 60$^{th}$ iteration of each synergy value, and a steady-state otherwise. In this work, this transient dynamics feature will be referred to as ``motor adaptation'', which represents an individual's improvement in performance over iterations of the task that eventually reaches a steady-state. As previously discussed, this is the behavior that introduces a challenge to the personalization of synergies due to the time (the number of iterations of the task execution) required to obtain the best performance that an individual can achieve using a given synergy setting.

This type of transient learning behavior has been extensively modeled for able-bodied individuals in the literature, with state-space models, exponentially decaying responses, and Linear Time Invariant (LTI) with output additive Gaussian white noise being widely accepted, e.g. \cite{Cheng2006, Wei2010, He2016, Zhou2012}. 
Although many models have been developed to characterize the dynamics of motor adaptation in able-bodied individuals and there is a growing body of work investigating the effects of prosthetic interfaces on motor adaptation \cite{Johnson2014a,Shehata2017}, a direct link between the dynamic models of able-bodied motor adaptation and the prosthetic case is yet to be established. 
It is worthwhile to highlight that the motor adaptation observed in Figure \ref{fig:sampleLTI} is captured by the increase in performance, as described by a given objective function, for a task by repetition, i.e. iteratively. The observed behavior suggests that a transient response in the iteration domain has a common structure and individual dependent parameters.

\subsection{Grey-box model for motor preference and adaptation}
\begin{figure}[ht]
    \centering
    %\vspace{-5mm}
    \includegraphics[trim={0.0in 0.0in 0.0in 0.0in}, width=3.0in, clip]{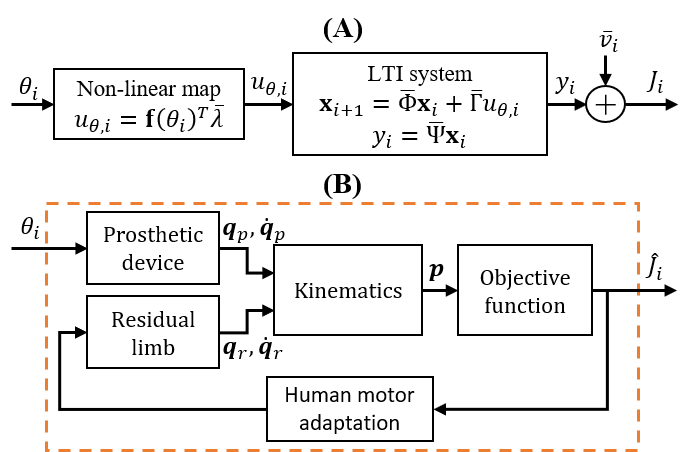}
    \vspace{-3mm}
    \caption{Overview of proposed human motor preference and adaptation model. (A) The grey-box model composed of a non-linear map and a LTI system in the iteration domain. (B) The block diagram of the human-prosthesis system represented by the model, as per the characterization of the task through an objective function ($\hat{J}_i$) and a parametrization of the prosthetic interface ($\theta_i$).}
    \label{fig:modelOverview}
    \vspace{-4mm}
\end{figure}

The proposed model for an individual's performance as a function of the synergy is constructed by the serial combination these two features: the average non-linear convex map (motor preference) and a perturbed LTI system in the iteration domain (motor adaptation). Figure \ref{fig:modelOverview}.(A) shows the diagram of the proposed model. Figure \ref{fig:modelOverview}.(B) shows the combined human-prosthesis system whose performance as a function of the synergy is being captured by the model.
Although there exist models that incorporate transient motor learning behavior in the iteration domain \cite{Zhou2012}, to the best of the authors' knowledge, a model that incorporates both preferences in motor behavior and individual transient learning behavior is novel in the literature.

The steady-state relationship between synergy and performance in the model represents the measured synergy-performance I-SSO map obtained experimentally as presented in Figure \ref{fig:sampleMaps}. The scalar $v_i$ characterizes human motor performance variation. $u_{\theta,i}\in\Real$ is the output of the average static map representing motor preference. One can think of $u_{\theta,i}$ as the best performance an individual can achieve for the specific synergy setting $\theta_i$, which occurs at the steady-state of motor learning. $y_i\in\Real$ is the output of the LTI system representing motor adaptation, which introduces transient dynamics to the average static map. This represents the internal cost, which is subsequently approximated using $\hat{J}_i$ from (\ref{eq:experimentalCost}).

%
% Nonlinear
%
\subsubsection{Motor preference}
\label{sec:nonlinearModel}
Given the average I-SSO relationship observed experimentally \cite{Garcia-Rosas2018EMBC}, motor preference is then modeled as a non-linear static input-output map. This map can be represented in general by a linearly-parameterized non-linear function
\begin{equation}
\label{eq:motorPreference}
    u_{\theta,i} = \fv(\theta_i)^{T}\bar{\lambdav},
\end{equation}
where $\fv(\cdot):\Theta\rightarrow\Real^{n_{\lambda}}$ is the vector of continuously differentiable characterizing functions, which is common across a population, and $\bar{\lambdav}\in\Real^{n_{\lambda}}$ is a constant vector of parameters, which is individual dependent and represents motor preference. 
In this case $u_{\theta,i}$ represents the average steady-state performance for a given $\theta_i$. The synergy $\theta_i$ is the current setting of the parameterization of the HPI and is considered as an input to the non-linear mapping. 
%\begin{remark}
%    $\fv(\cdot)$ is considered to be a common basis across a broad population, whilst the individual preferences are captured in the parameter vector $\bar{\lambdav}$. \hfill $\circ$
%\end{remark}

The following assumption is made regarding the characterization of motor preference.
\begin{assumption}
\label{as:mapConvexity}
 There exists a unique maximum $\theta^{*}\in\Theta$ such that for any given $\bar{\lambdav}\in\Real^{n_{\lambda}}$, there is $u_{\theta}^* = \fv(\theta^{*})^{T}\bar{\lambdav}$, $\theta^{*}\in\Theta$, and 
    \begin{subequations}
    \begin{eqnarray}
        u^{'}_\theta(\theta) = 0, & \textup{iff} \quad \theta = \theta^{*},\\
        \label{eq:curvature}
        u^{''}_\theta(\theta) < 0, & \forall \theta\in \Theta.
    \end{eqnarray}
    \end{subequations}
\end{assumption}
\begin{remark}
    Assumption \ref{as:mapConvexity} can be satisfied by carefully formulating the interface parametrization and measure of task performance, as previously presented in \cite{Garcia-Rosas2018EMBC}. Although $u_\theta^*$ depends on the parameter $\bar{\lambdav}$, the optimal parameter $\theta^*$ is independent of the parameter $\bar{\lambdav}$. From experimental results \cite{Garcia-Rosas2018EMBC}, it was observed that individuals who reached steady-state task performance with a virtual synergistic prosthetic device satisfy this assumption locally, as seen in Figure \ref{fig:sampleMaps}. \hfill $\circ$
\end{remark}

% LTI
%
\subsubsection{Motor adaptation}
\label{sec:linearModel}
A model to characterize the transient behavior observed in performance ($J$) as the synergy setting ($\theta$) changed from one value to the next, which can be attributed to motor adaptation, is presented next. In this work, due to repetitive nature of the tasks being considered, a Linear Time-Invariant (LTI) state-space model in the iteration domain is utilized to represent human learning behavior. Human motor variation is represented as additive noise with respect to the given performance. Such representation is natural to represent the experimental and application setting as performance over one iteration of the task is used as an output. At the $i^{th}$ iteration, the proposed model takes the following form:
\begin{subequations}
%\vspace{-5mm}
\label{eq:LTImodel}
\begin{align}
    \xv_{i+1} &= \bar{\Phi}\xv_{i} + \bar{\Gamma} u_{\theta,i} \\
    J_{i} &= \bar{\Psi}\xv_{i} + \bar{v}_{i},
%\vspace{-3mm}
\end{align}
\end{subequations}
where $\xv\in\Real^{n_h}$ represents the internal states of human motor adaptation and $n_{h}>0$ is the order of the model. The matrices $\bar{\Phi}$, $\bar{\Gamma}$, and $\bar{\Psi}$ have the appropriate dimensions and are individual dependent. Performance is perturbed by human motor variation, which is represented by the output additive noise $\bar{v}$ \cite{Cheng2006}.
An example of the observed motor adaptation behavior is presented in Figure \ref{fig:sampleLTI}. The following assumption is utilized:
\begin{assumption}
\label{as:steadyStateGain}
    System (\ref{eq:LTImodel}) has unity steady-state gain. \hfill
    ($G_{ss} := \bar{\Psi}(I_n-\bar{\Phi})^{-1}\bar{\Gamma} = 1$). \hfill $\circ$
\end{assumption}
\begin{remark}
    Assumption \ref{as:steadyStateGain} allows the map $u_{\theta,i}$ (\ref{eq:motorPreference}) to represent an individual's average static relationship between synergy ($\theta$) and performance ($J$). Furthermore, at steady-state $y_{ss}$ approaches $u_{\theta,ss}$. This can be readily satisfied through the selection of $\bar{\lambdav}$. \hfill $\circ$
\end{remark}
This concludes the proposed model structure to represent human motor preference and adaptation. It is important to highlight that the parameters of this model are individual dependent. Next, the identification of the parameters for two individuals, used for validation, is presented.

%
% Simulation results
%
%
%\vspace{-3mm}
\subsection{Model parameter identification}
In order to validate the proposed model structure, system identification was performed on synergy-performance data for two representative able-bodied subjects. Transhumeral amputation and prosthesis use was emulated through a Virtual Reality Environment (details in Section \ref{sec:experiment}). The simulation of the identified systems was compared to a different synergy-performance data-set from these individuals.
The identification procedure is as follows: first, the data collected is used to identify the parameters of the non-linear I-SSO map (motor preference), followed by the identification of the normalized LTI system (motor adaptation). Finally, the characteristics of motor performance variation ($\bar{v}_i$) are obtained from the analysis of residuals between the validation data and the identified underlying model.

The dataset for the two individuals used for system identification was taken from the authors' previous work in \cite{Garcia-Rosas2018EMBC} and a new dataset from the same individuals was generated for comparison with the identified model and validation. The original dataset was generated by holding a synergy constant over multiple iterations of a forward reaching task with a synergistic prosthetic elbow. This procedure was followed in order to capture both the transient and steady-state behavior for each synergy. This was repeated for multiple synergies to obtain the non-linear map. The validation dataset was generated with the same task and subjects; however, the synergy was changed linearly in every task iteration instead; capturing both features in the response.

%
% Map ID
%
\subsubsection{Motor preference}
\begin{figure}[ht]
    \centering
    \vspace{-3mm}
    \includegraphics[trim={0.0in 0.0in 0.0in 0.0in}, width=3.4in, clip]{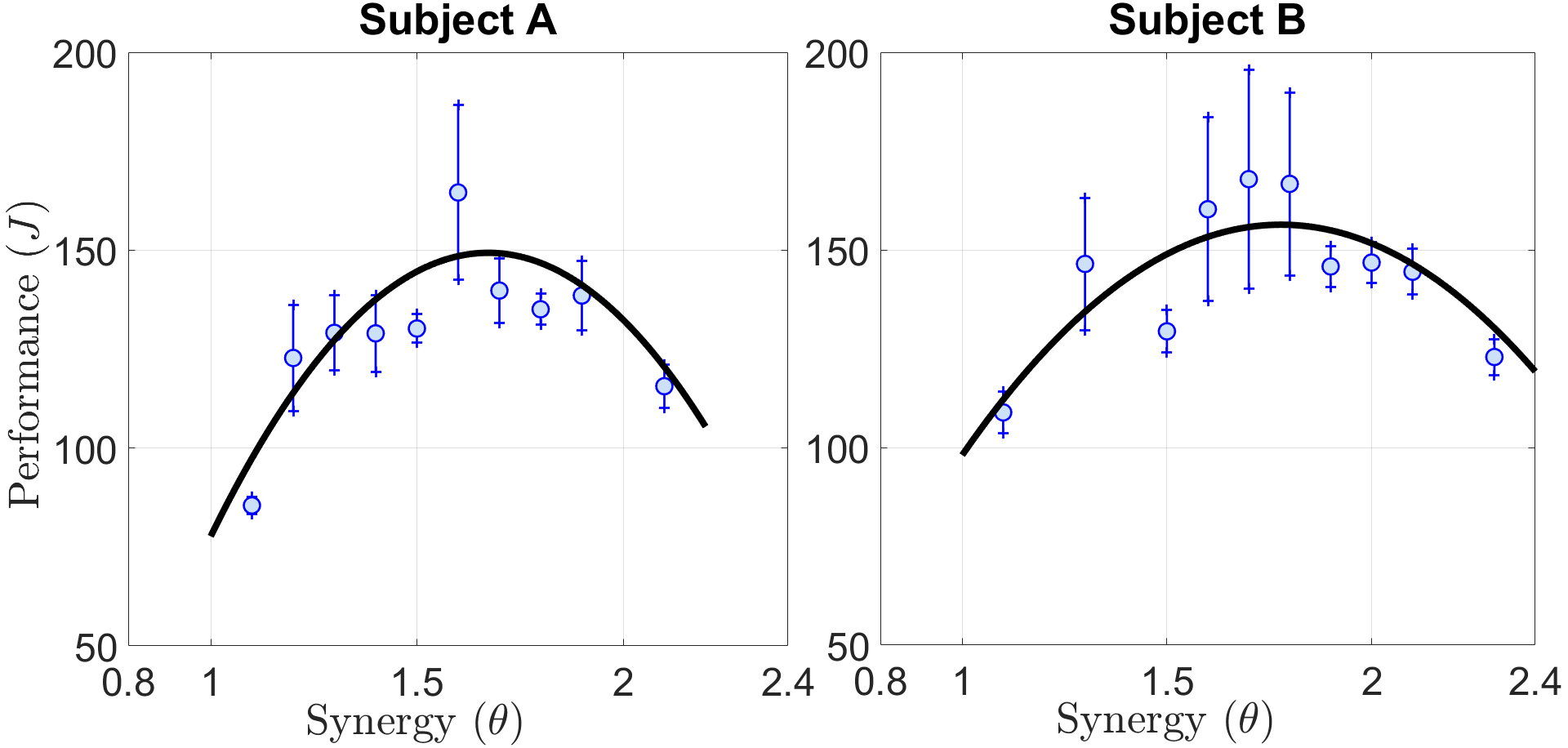}
    \vspace{-3mm}
    \caption{Non-linear map observed experimentally with quadratic fit for two representative subjects. The blue circles represent mean performance, the whiskers the standard deviation, and the black line the quadratic polynomial representing the identified map.}
    \label{fig:identifiedNonlinear}
    \vspace{-1mm}
\end{figure}

In order to identify the average I-SSO map representing motor preference, the structure of $\fv(\theta)$ is needed. As in \cite{Garcia-Rosas2018EMBC}, the nonlinear mapping was approximated as a quadratic polynomial. The linearly parameterized non-linear function representing motor preference is given by
\begin{equation}
\label{eq:quadMap}
    J(\theta)=u_{\theta,ss}(\theta) = \fv(\theta)^{T}{\bar \lambdav} = \begin{bmatrix}
    \theta^2 & \theta & 1
    \end{bmatrix}{\bar \lambdav},
\end{equation}
with $\bar{\lambdav}\in\Real^{3}$ being the parameters identified for each individual. The map was identified from experimental data using a least squares method. Figure \ref{fig:identifiedNonlinear} shows the identified $J(\theta)$ map for two subjects using a quadratic polynomial along with the experimental data used for identification. The identified map parameters $\bar{\lambdav}$ for both subjects are the following.

\begin{subequations}
\vspace{-4mm}
\begin{align}
    \bar{\lambdav}_{A} &= \begin{bmatrix}
    -158.15 & 529.18 & -293.34
    \end{bmatrix}^{T},
    \label{eq:mapA} \\
    \bar{\lambdav}_{B} &= \begin{bmatrix}
    -96.18 & 342.13 & -147.86
    \end{bmatrix}^{T}.
    \label{eq:mapB}
\end{align}
\end{subequations}

\subsubsection{Motor adaptation}
There are two components in (\ref{eq:LTImodel}) that need to be identified in order to characterize an individual's learning dynamics: the underlying LTI model and the associated noise ($\bar{v}$). Undershoot in performance has been observed to occur following step synergy changes, as illustrated in Figure \ref{fig:sampleLTI}. This can be attributed to the individual's reduced performance due to the sudden change in the synergy setting applied and the gradual re-adaptation to the dynamics of the new synergy setting.
Thus, systems of order greater than one are considered for identification, as these are the minimum required to capture undershoot.
Furthermore, as overshoot is non-physical in the timescales considered, constrained least squares was used for identification in order to limit the identified systems to be over-damped. The identified non-linear I-SSO map from the previous section was used as the input $u_{\theta}$ to the LTI subsystem.

The experimental data for the transient response corresponding to a single synergy value was used for validation of the identified LTI system. Validation was performed through the comparison of the Mean Square Error (MSE) between the response of the identified underlying model and the validation data-set for different system orders \cite{Landau2011}. 
%Figure \ref{fig:LTIcomp} presents the validation data along with the response of the identified underlying model. 
Table \ref{ta:MSE} contains the resultant MSE for the second and third order cases for both subjects.

\begin{table}[ht]
    %\vspace{-3mm}
    \caption{Mean Square Error (MSE) for the second and third order LTI system models.}
    \vspace{-2mm}
    \centering
    \begin{tabular}{|c||c|c|}
        \hline
        Subject & 2nd order MSE & 3rd Order MSE  \\
        \hline
        \hline
        A       &     277.76     &    274.36      \\
        \hline
        B       &     489.93    &     467.37       \\
        \hline
    \end{tabular}
    \label{ta:MSE}
    \vspace{-4mm}
\end{table}

As seen in the validation results presented in Table \ref{ta:MSE}, both the second and third order models achieve similar MSE. Since it is desirable to use the lowest complexity model, and the second order model captures the key transient behavior, the second order model is selected. The identified normalized LTI system matrices representing the underlying individual transient behavior for subjects A and B are listed below.
% CS eqns
\begin{subequations}
\label{eq:LTIA}
\begin{eqnarray}
    \bar{\Phi}_{A} &=& \begin{bmatrix}
    0 & 1 \\
    0.068 & 0.35
    \end{bmatrix},
    \\
    \bar{\Gamma}_{A} &=& \begin{bmatrix}
    0.839 &
    0.037
    \end{bmatrix}^T, 
    \bar{\Psi}_{A} = \begin{bmatrix}
    1 & 0
    \end{bmatrix},
\end{eqnarray}
\end{subequations}
% HT eqns
\begin{subequations}
\vspace{-5mm}
\label{eq:LTIB}
\begin{eqnarray}
    \bar{\Phi}_{B} &=& \begin{bmatrix}
    0 & 1 \\
    -0.017 & 0.25
    \end{bmatrix},
    \\
    \bar{\Gamma}_{B} &=& \begin{bmatrix}
    -0.091 &
    0.834
    \end{bmatrix}^T, 
    \bar{\Psi}_{B} = \begin{bmatrix}
    1 & 0
    \end{bmatrix}.
\end{eqnarray}
\end{subequations}

Finally, an analysis of the residuals between the validation data and the identified LTI system was performed in order to obtain the characteristics of motor performance variation ($\bar{v}_i$). This was done by performing a whiteness (independence) test of the residuals in order to establish whether $\bar{v}_i$ can be characterized by Gaussian white noise, and the mean and variance characteristics for each subject \cite{Landau2011}. The whiteness test was performed with a $95\%$ confidence interval and the validation data-set for each subject contained 50 samples, such that the validation criterion was $0.277$. This criterion represents the maximum auto-correlation permitted between the residuals in order to accept the hypothesis that the residuals have a Gaussian distribution. The resultant normalized auto-correlations from the analysis of residuals were $0.1495$ and $0.227$ for subject A and B respectively, both smaller than the validation criterion. Therefore, it is concluded with $95\%$ confidence that $\bar{v}_i$ has a Gaussian distribution. The individual $\bar{v}_i$ characteristics for each subject are presented in Table \ref{ta:noiseCharac}.

\begin{table}[ht]
\vspace{-3mm}
    \caption{Mean and standard deviation of residuals obtained from whiteness test.}
    \vspace{-2mm}
    \centering
    \begin{tabular}{|c||c|c|}
        \hline
        Subject & Mean ($\mu$) & Standard deviation ($\sigma$)  \\
        \hline
        \hline
        A       &     0     &     16.81      \\
        \hline
        B       &     0     &     22.36        \\
        \hline
    \end{tabular}
    \label{ta:noiseCharac}
    \vspace{-5mm}
\end{table}

%
% Simulation
%
\subsection{Model validation}
\label{sec:modelValidation}
\begin{figure}[ht]
    \centering
    \vspace{-2mm}
    \includegraphics[trim={0.0in 0.0in 0.0in 0.0in}, width=3.4in, clip]{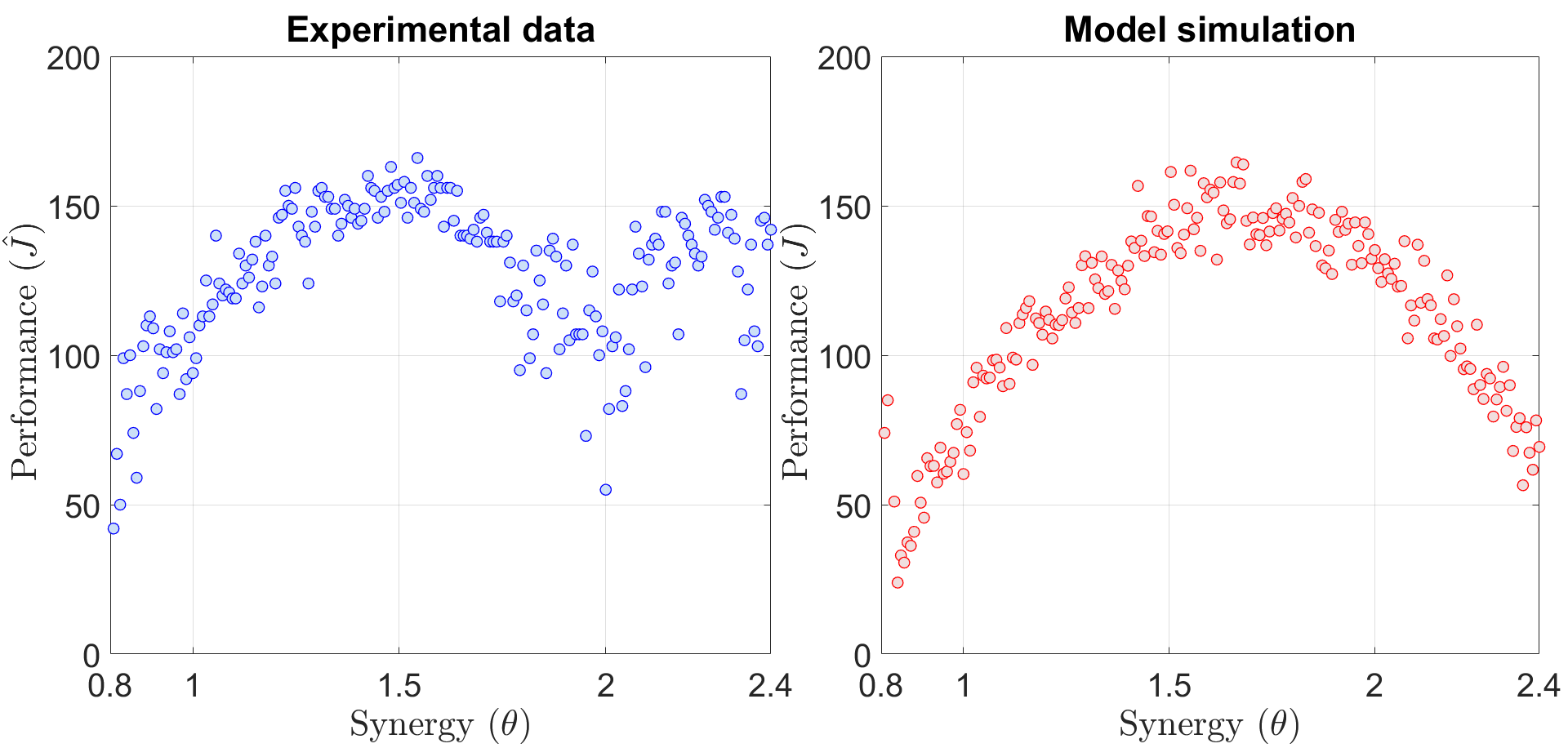}
    \vspace{-3mm}
    \caption{Subject A experimental and model simulation synergy sweep performance response.}
    \label{fig:sweep_A}
    \vspace{-6mm}
\end{figure}

\begin{figure}[ht]
    \centering
    %\vspace{-0mm}
    %\includegraphics[trim={0.0in 0.0in 0.0in 0.0in}, width=3.4in, clip]{Figures/sweepHT.eps}
    \includegraphics[trim={0.0in 0.0in 0.0in 0.0in}, width=3.4in, clip]{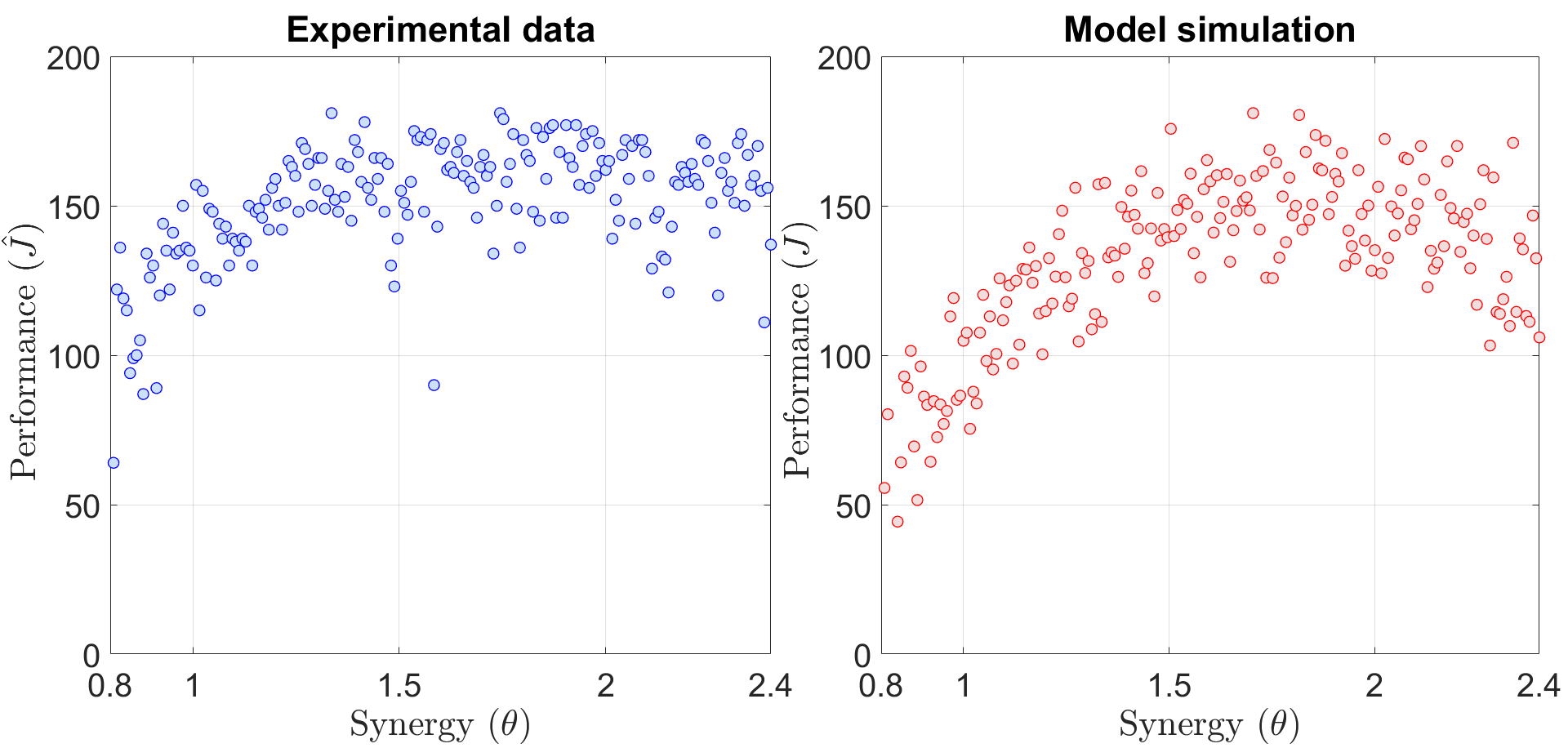}
    \vspace{-3mm}
    \caption{Subject B experimental and model simulation synergy sweep performance response.}
    \label{fig:sweep_B}
    \vspace{-2mm}
\end{figure}

The identified model of the individual subject behavior (subject A and subject B above) is numerically simulated and the output was compared to an independently obtained validation data-set for the respective subjects. The experimental data-set for validation was generated for the same forward reaching task as the identification data-set. However, in this case the synergy was changed at each iteration according to:

\vspace{-2mm}
\begin{equation}
    \theta_i = 0.8 + \frac{i}{125},
\end{equation}
where $i$ represents the $i$th iteration of the task. The synergy range studied, $\theta_{min} = 0.8$ and $\theta_{max} = 2.4$, was informed by preliminary results in \cite{Garcia-Rosas2018EMBC}. Performance of the task for a given iteration was measured by the objective function (\ref{eq:experimentalCost}). Figures \ref{fig:sweep_A} and \ref{fig:sweep_B} present the experimental data and simulation results for subjects A and B respectively.


It can be seen from Figures \ref{fig:sweep_A} and \ref{fig:sweep_B} that the structure of the model is able to qualitatively capture the common features of motor preference (the shape of the non-linear I-SSO map) and adaption (the transient behavior) across the population, as well as human variation given the identified parameters $\bar{\lambdav},\bar{\Phi},\bar{\Gamma},\bar{\Psi},\bar{v}$ depend on individuals. In the validation data for subject A, presented in Figure \ref{fig:sweep_A}, an increase in performance can be observed for $\theta >2$. This behavior may be attributed to the subject changing their motor strategy in order to improve their performance for that synergy range. It was anecdotaly observed, however, that this change in strategy relied on upper body and shoulder compensation motion. This highlights a limitation of the chosen measure of performance (\ref{eq:experimentalCost}), as it does not capture compensation motion or metabolic effort. Nonetheless, even with such diverse human variation, both individually and across the population, the model adequately captures both the underlying common behavior of a population and individual behavior.

With the existence of a personalized interface parameter optimal value and the common model structure identified from experimental results, a grey-box based on-line optimization technique \cite{Moase2011a, Moase2009a} can be naturally linked to the kinematic synergies personalization problem in HPIs. This technique is able to find the optimal personalized solution to the interface parameter to performance problem in the presence of human variation.
Differing from black-box on-line optimization techniques with personalized tuning of parameters and personalized convergence performance, the grey-box on-line optimization technique utilizes the common model structure to obtain a uniformly convergent performance among the population with tuning parameters that are less sensitive to human variation.

%
% Human-Robot Interface personalization algorithm 
%
\section{Human-Prosthesis Interface personalization algorithm}
\label{sec:algorithm}
%
% Description
%
The on-line personalization algorithm is presented in detail in this section, followed by simulation results using the proposed grey-box model. Simulations were done to tune the algorithm, test its behavior, and observe its performance. The algorithm takes advantage of the previously modeled structure of human motor preference and learning by allowing it to estimate the steady-state performance from on-line measurements for the given synergy value regardless of the stage of learning (transient behavior). Furthermore, it is able to extract additional information from the estimates of the non-linear map to adjust its update rate to accommodate individual motor behavior characteristics. The algorithm is adapted from \cite{Moase2011a} and \cite{Moase2009a} for the HPI application. The major change is that the current setting is in the iteration domain due to the event-driven iterative nature of the problem.

\begin{figure}[ht]
    \centering
    \vspace{-2mm}
    \includegraphics[trim={0.0in 0.0in 0.0in 0.0in}, width=3.4in, clip]{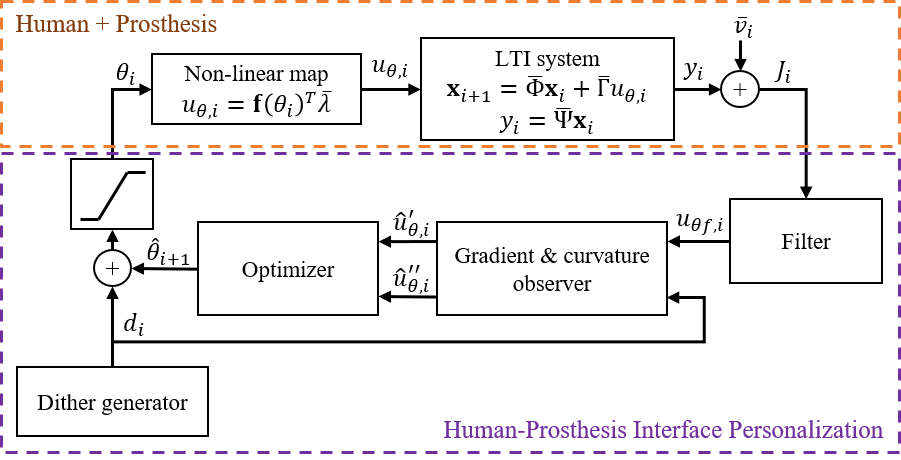}
    \vspace{-3mm}
    \caption{General overview of the proposed framework for personalization of human-prosthesis interfaces. }
    \label{fig:frameworkDiagram}
    \vspace{-2mm}
\end{figure}

Figure \ref{fig:frameworkDiagram} presents the diagram of the proposed framework for personalization of human-prosthesis interfaces. The framework is composed of four different components: a filter, a gradient and curvature observer, an optimizer, and a dither generator; which will be detailed next. This particular structure is designed to operate under the conditions satisfied by the model of human motor preference and adaptation.

The algorithms in \cite{Moase2011a} and \cite{Moase2009a} require that certain conditions are satisfied to ensure that the parameter $\theta$ will converge close to the optimal $\theta^*$. A summary of these conditions is presented herein.
\begin{condition}
\label{co:extremum}
    There exists an extremum $\theta^{*}$ in the non-linear steady-state map $J(\theta)$ and the map is twice continuously differentiable \cite[Assumption 2]{Moase2011a}.
\end{condition}
\begin{condition}
\label{co:degree}
    The relative degree of the LTI dynamic system is known \cite[Condition 6]{Moase2011a}.
\end{condition}
\begin{condition}
\label{co:minPhase}
    The serial combination of the LTI System and the Filter (Figure \ref{fig:frameworkDiagram}) is minimum phase \cite[Condition 6]{Moase2011a}.
\end{condition}
\begin{condition}
\label{co:PE}
    Persistency of excitation (PE) is satisfied for parameter estimation.
\end{condition}

Assumption \ref{as:mapConvexity} ensures that Condition \ref{co:extremum} is satisfied. Section \ref{sec:model} identified that the normalized LTI system has a relative degree one, thus Condition \ref{co:degree} is satisfied. Given that the system (\ref{eq:LTImodel}) represents human motor adaptation behavior, it indicates that the matrix $\Phi$ is stable, thus satisfying Condition \ref{co:minPhase}. 
Persistency of excitation is a common condition required in adaptive systems. Condition \ref{co:PE} can be satisfied by introducing a known dither (perturbation) signal that excites the parameters (synergy setting) that need to be identified.
%

%
% Filter
%
\subsection{Filter} 
The role of the filter is to extract the information of the motor performance non-linear map $u_{\theta}$ from the measured task performance $J$, which is subject to learning dynamics and contains performance variability and measurement noise. It is the component that ensures that the tuning parameters are independent of individual motor adaptation dynamics.

Conditions \ref{co:degree} and \ref{co:minPhase} are needed for filter design since the objective of the filter is to invert the dynamics of the LTI system. Ideally, this allows the estimation of the non-linear map to be insensitive to the individual's learning behavior. With these conditions satisfied, then the stability of the closed loop can be ensured and an appropriate filter for the LTI system can be designed. In this work, a band-pass filter is used in order to maintain causality. Let the dither signal frequency range be $[\omega_o \quad n\omega_o]$, then the band-pass filter can be represented in state-space as follows.
\begin{subequations}
\begin{align}
    \wv_{i+1} &= \Phi_{f}(\omega_o,n\omega_o)\wv_{i} + \Gamma_{f} J_{i}, \\
    u_{\theta f,i} &= \Psi_{f}(\omega_o,n\omega_o)\wv_{i},
\end{align}
\end{subequations}
where $\wv\in\Real^{n_{f}}$ is the state of the filter. The order of the filter $n_{f} > 0$ is dependent on the LTI system. As the LTI system identified from the experiments has relative degree one, a second order filter is used with $n_f=2$. Hence, the matrix $\Phi_{f}(\omega_o,n\omega_o)$ is in $\Real^{2\times 2}$, the matrix  $\Gamma_{f}$ in $\Real^{2}$, and the matrix $\Psi_{f}(\omega_o,n\omega_o)$ in $\Real^{1\times 2}$. 
%
% Gradient estimator
%
\subsection{Gradient and Curvature Observer} 
\label{sec:observer}
This component estimates the gradient and curvature information from the motor performance non-linear map $u_{\theta}$ that is used in the optimizer to find the optimal individual synergy $\theta^*$, provided that Condition \ref{co:PE} is satisfied. There are multiple methods for performing these estimates, such as demodulation \cite{Nesic2010a} and observer based \cite{Moase2011a}.

In this work, a discrete version of the Luenberger observer method presented in \cite{Moase2009a} is used. The observer takes the following form 
\begin{subequations}
\begin{eqnarray}
    \hat{u}_{\theta f, i} &=& \Psi_o \zvh_{i} \\
    \zvh_{i+1} &=& \omega_{o}\Phi_{o}\zvh_{i} + \omega_{o}\Lv (u_{\theta f, i} - \hat{u}_{\theta f, i})
\end{eqnarray}
\end{subequations}
where $\Lv\in\Real^{5}$ is the observer gain which is designed such that $\Phi_{o}-\Lv\Psi_{o}$ is stable, so that the estimated gradient and curvature will converge to the true values. The matrices of the observer take the following form as in \cite{Moase2009a}:
\begin{equation}
    \Phi_{o} = \begin{bmatrix}
    0 & 0  & 0 & 0  & 0\\
    0 & 0  & 1 & 0  & 0\\
    0 & -1 & 0 & 0  & 0\\
    0 & 0  & 0 & 0  & 2\\
    0 & 0  & 0 & -2 & 0
    \end{bmatrix},
    \Psi_{o} = \begin{bmatrix}
    1 \\
    1 \\
    0 \\
    0 \\
    -0.25
    \end{bmatrix}^T.
\end{equation}

Then $u^{'}_{\theta}$ and $u^{''}_{\theta}$ can be estimated by demodulating the states $\zvh$, such that
\begin{subequations}
\begin{align}
    \hat{u}^{'}_{\theta, i} &= \frac{1}{a}\Psi^{'}\zvh_{i}, & \Psi^{'} &= \begin{bmatrix}0 & \mcl{S}_{1} & \mcl{C}_{1} & 0 & 0\end{bmatrix}, \\
    \hat{u}^{''}_{\theta,i} &= \frac{1}{a^{2}}\Psi^{''}\zvh_{i}, & \Psi^{''} &= \begin{bmatrix}0 & 0 & 0 & \mcl{S}_{2} & \mcl{C}_{2} \end{bmatrix},
\end{align}
\end{subequations}
where $\mcl{S}_{n} = \sin{(n\omega_{o} i)}$ and $\mcl{C}_{n} = \cos{(n\omega_{o} i)}, n=1,2$ are the dithers needed for estimation, and $i>0$ is the iteration index. By using $\mcl{S}_{n}$, $\mcl{C}_{n}$, and standard averaging techniques \cite[Chapter 10]{Khalil2002}, the output of the observer will converge close to the gradient and curvature of the non-linear map.

\begin{remark} 
    Curvature estimation is on a faster time-scale than that of gradient estimation due to Condition \ref{co:PE}. Hence the demodulation dither signals $\mcl{S}_{2}$ and $\mcl{C}_{2}$ operate on a $2\omega_{o}$ frequency. Given the iterative nature of the system, care must be taken when selecting $\omega_{o}$ to avoid issues such as aliasing.\hfill $\circ$
\end{remark}

%
% Optimizer
%
\subsection{Optimizer} 
\label{sec:optimizer}
Once the gradient and curvature of the non-linear map are estimated, they can be used in an optimizer to find the optimal $\theta^*$. With consideration of human variations, the optimizer is designed such that it can achieve uniform convergence performance, which is independent to the shape of the non-linear map $u_\theta(\cdot)$, i.e. operates similarly on individuals with different motor behavior characteristics.

A widely used optimizer is gradient-ascent, see for example \cite{Ariyur2003, Garcia-Rosas2018ACC} and references therein. The gradient-ascent method with a fixed step size is simple to implement, but the convergence speed is determined by the shape of the non-linear map (given by $\bar{\lambda}$ in this work). In order to achieve uniform convergence speed across individuals, the derivative based optimizer employed in this paper switches between a gradient-ascent and a Newton-like approach. The reason to add the Newton-like approach is two-fold. First, the faster convergence speed. Second, its performance being less sensitive to the shape of the non-linear map.
Although, the shape of the non-linear map varies over the population, convergence speed of the proposed HPI personalization algorithm is insensitive to this variation. 

Assumption \ref{as:mapConvexity}, particularly equation (\ref{eq:curvature}), is  needed to ensure that the Newton-like optimizer is applicable. However, this can only be satisfied locally as shown from experimental results in \cite{Garcia-Rosas2018EMBC}. This local region is usually smaller than the region where gradient-descent can converge. 
Thus the Newton-like optimizer must operate within the region in which equation \ref{eq:curvature} is satisfied. The gradient-descent optimizer operates outside that region. As such, the following switching mechanism is used.
\begin{subequations}
\begin{align}
    \hat{\theta}_{i+1} &= \hat{\theta}_{i} + k\omega_{o}\Delta_{i} \\
    \Delta_{i} &= \left\{\begin{matrix} 
    - \frac{\hat{u}^{'}_{\theta,i}}{\hat{u}^{''}_{\theta,i}} & \|\hat{u}^{'}_{\theta,i}\| < -\varepsilon\hat{u}^{''}_{\theta,i},\\
    \hat{u}^{'}_{\theta,i} & \textup{otherwise},
    \end{matrix}\right.
\end{align}
\end{subequations}
where $\varepsilon > 0$ determines the size of the region where a Newton step is employed, and $k>0$ is the optimizer gain. The condition $\|\hat{u}^{'}_{\theta,i}\| < -\varepsilon\hat{u}^{''}_{\theta,i}$ serves two purposes. First, to ensure that when the Newton-like optimizer is used, equation \ref{eq:curvature} of Assumption \ref{as:mapConvexity} is satisfied. Second, to avoid numerical issues arising when $\hat{u}^{''}_{\theta,i}$ is close to zero. 
\begin{remark}
    An important consideration with the tuning of this algorithm is the time-scale separation between the observer and the synergy ($\theta$) rate of change. 
    The change in $\theta$ needs to be sufficiently slow to ensure that the observer provides relatively accurate estimation of the gradient and curvature. By carefully tuning the optimizer gain ($k$), such time-scale separation can be achieved. The tuning guideline can be found in \cite{Moase2009a}.\hfill $\circ$
\end{remark}

%
% Dither Generator
%
\subsection{Dither Generator}
\label{sec:dither}
In order to be able to estimate the gradient and curvature of the non-linear map $u_\theta(\cdot)$, as a standard requirement in parameter identification, Condition \ref{co:PE} is  needed. This is done by perturbing the system with a known dither vector \cite{Moase2009a}. Since gradient and curvature need to be estimated, two different dither signals are required.

Let the estimation frequency to be used for gradient be $\omega_o$. In this work, a sinusoidal dither generator is used, leading to the following dither vector: 
\begin{equation}
    \dv = \begin{bmatrix}
    a\mcl{S}_{1} & a\mcl{S}_{2}
    \end{bmatrix}^T
\end{equation}
where $a>0$ is the dither amplitude.
\begin{remark}
\label{re:ditherVar}
    It has been found that high variability can impair human motor learning \cite{Cardis2017}, so care must be taken to ensure the dither amplitude is not too large.\hfill $\circ$
\end{remark}

%
%
% Simulation results
%
%
\subsection{Personalization algorithm implementation with simulated humans}
With the multiple conditions that need to be satisfied for the application and tuning of the algorithm, it is desirable to test it on simulated humans before it is applied to human-in-the-loop applications. This simulation can be performed using the proposed model and identified parameters presented in the previous section. Here the kinematic synergies personalization simulation results for the two subjects previously modeled are presented.

Tuning of the algorithm was performed by first selecting the dither frequency $w_{o}$ and amplitude $a$ as they play an important role to ensure convergence. Considering the application, it is desirable to achieve convergence of algorithm within the time for a prosthesis fitting and training session. From previous experiments with reaching tasks, it was found that 100 to 200 iterations are acceptable for a training session as this number of iterations take 30 to 60 minutes to perform. In the sequel, the dither frequency needed is selected as $\omega_{o} = \pi/4$. Given that the system is of relative degree one, a second order band-pass Q-filter was chosen. The following parameters were used for its design: $H = 0.5$ and $Q = 5.0$.

The choice of $a$ presents one of the first challenges for a human-in-the-loop system (see Remark \ref{re:ditherVar}). From the point of view of the human, it would be difficult to learn to use a device that is constantly and unpredictably changing its behavior. Furthermore, the choice of $a$ will affect the estimator. From previous experiments, it was found that $a<=0.05$ is well received by subjects, thus a dither amplitude of $a = 0.02$ was chosen. Finally, the optimizer parameters were chosen by evaluating the system response with a range of optimizer gain $k$ values. Final algorithm parameters were: observer gain $\Lv = \begin{bmatrix} 1.5 & 0.25 & 0.25 & 2.0 & -2.0 \end{bmatrix}$, $\varepsilon = 0.1$, and $k = 0.05$. 

\begin{figure}[ht]
    \centering
    \vspace{-2mm}
    \includegraphics[trim={0.0in 0.0in 0.0in 0.0in}, width=3.4in, clip]{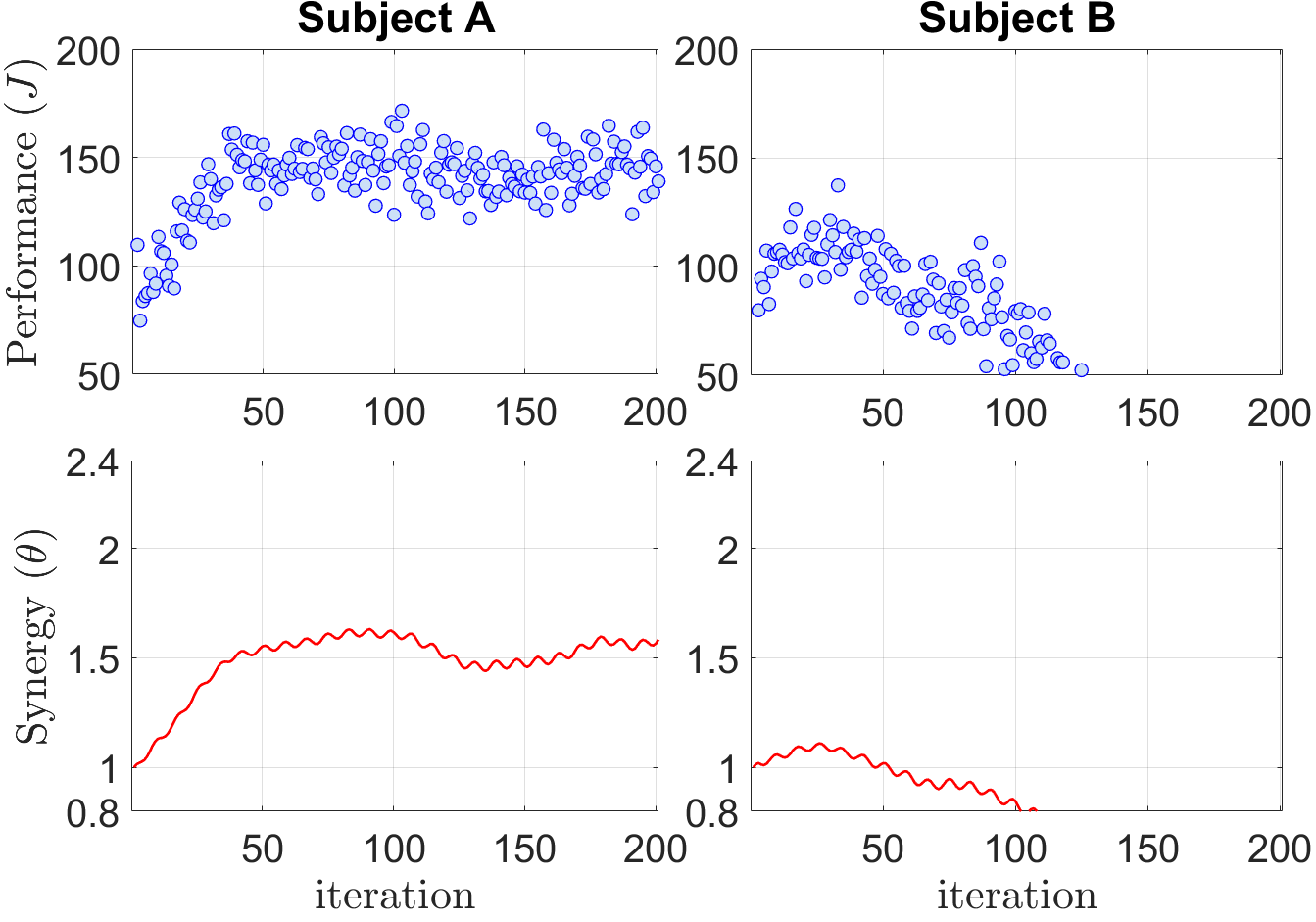}
    \vspace{-4mm}
    \caption{Simulation results for the previous algorithm (\cite{Garcia-Rosas2018ACC}) for the two modelled subjects. Top plots show performance across iterations while bottom plots show synergy across iterations. The algorithm fails to identify the optimal synergy for subject B which leads to the subject not being able to achieve the task accurately and thus diminishing performance.}
    \label{fig:simCounterExample}
    \vspace{-2mm}
\end{figure}

The algorithm the authors previously presented in \cite{Garcia-Rosas2018ACC} was also implemented. Such algorithm uses a ``black-box'' on-line optimization technique without exploring the transient behavior of human learning. Hence it suffers the disadvantage that the tuning parameters are sensitive to person-to-person variation. In order to demonstrate such sensitivity, the same tuning parameters are used for both individuals. Algorithm parameters $a$ and $w$ were equivalent to those used for the proposed algorithm, while $k$ was modified to $k = 0.005$ as required by the different optimizer. Simulation results for the algorithm in \cite{Garcia-Rosas2018ACC} are presented in Figure \ref{fig:simCounterExample}, where performance and synergy for the two subjects is shown.

It is clear in Figure \ref{fig:simCounterExample} from the decreasing performance of subject B that the previous algorithm \cite{Garcia-Rosas2018ACC} failed to identify the synergy for this subject when the same algorithm parameter tuning was used for both subjects. The decrease in performance implies that the subject was not able to perform the task fast and accurately with the given synergy values, as defined by the objective function. Practically, this means that the elbow does not extend sufficiently for the given shoulder flexion and thus the subject cannot reach adequately. The failure to identify the optimal synergy is due to the tuning parameters of this algorithm being dependent on each individual.

\begin{figure}[ht]
    \centering
    %\vspace{-2mm}
    %\includegraphics[trim={0.0in 0.0in 0.0in 0.0in}, width=3.4in, clip]{Figures/simResults.eps}
    \includegraphics[trim={0.0in 0.0in 0.0in 0.0in}, width=3.4in, clip]{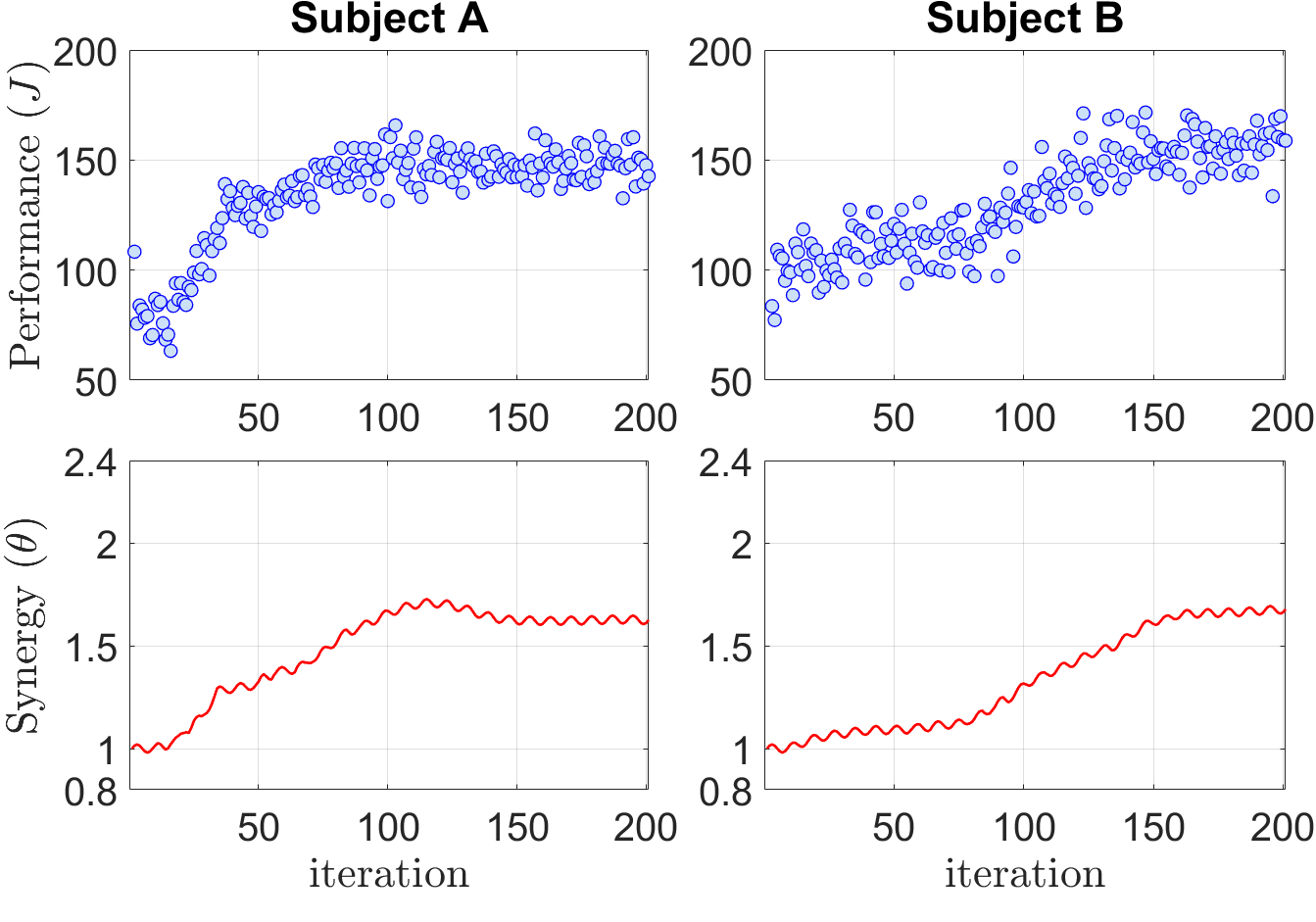}
    \vspace{-5mm}
    \caption{Proposed algorithm simulation results for performance and synergy for the two modelled subjects. The top plots show performance across iterations while the bottom plots show synergy across iterations. The algorithm successfully identifies the optimal synergy for both subjects.}
    \label{fig:simFESResults}
    \vspace{-2mm}
\end{figure}

Simulation results for the algorithm proposed in this paper are presented in Figure \ref{fig:simFESResults}, where performance $(J)$ and synergy $(\theta)$ of the two modeled subjects is shown.
As can be observed in Fig. \ref{fig:simFESResults}, the proposed algorithm updates the synergy over iterations of the task until the maximum performance (as defined by the identified map used for simulation) is reached. Thus, the proposed algorithm was able to successfully identify the synergy for the two modeled subjects with different motor behavior characteristics while using the same tuning parameters. 
This demonstrates the capability of the proposed algorithm to handle individual motor preference and adaptation dynamics without the need for individual tuning.

\section{Synergistic prosthesis on-line personalization experiment}
\label{sec:experiment}
A human-in-the-loop experiment in a Virtual Reality Environment (VRE) was designed to emulate upper limb loss and prosthesis use in able-bodied subjects. The objective of the experiment was to experimentally evaluate the effectiveness of the proposed algorithm to realize on-line personalization for a forward reaching task. Furthermore, to demonstrate that the proposed algorithm does not require the individual tuning for a population of subjects with different motor behavior and learning characteristics.

For ease of presentation, a simple point-to-point forward reaching task and scalar linear synergy between shoulder and elbow flexion/extension were chosen, where shoulder flexion translated to elbow extension. This scalar synergy, however, has been found to be able to represent the most significant relationship between shoulder and elbow for forward reaching tasks \cite{Bockemuhl2010}; and has been successfully applied to synergistic prosthetic elbow studies \cite{Alshammary2018}. It should be noted that the general case method proposed herein is designed to accommodate a wide variety of tasks through the choice of objective function, and more complex synergy parametrizations and structures due to its black box approach to parameter identification.

%
% Description
%
\subsection{Experiment description}
% Platform setup
\begin{figure*}[ht!]
    \centering
    %\vspace{-5mm}
    \includegraphics[width=6in]{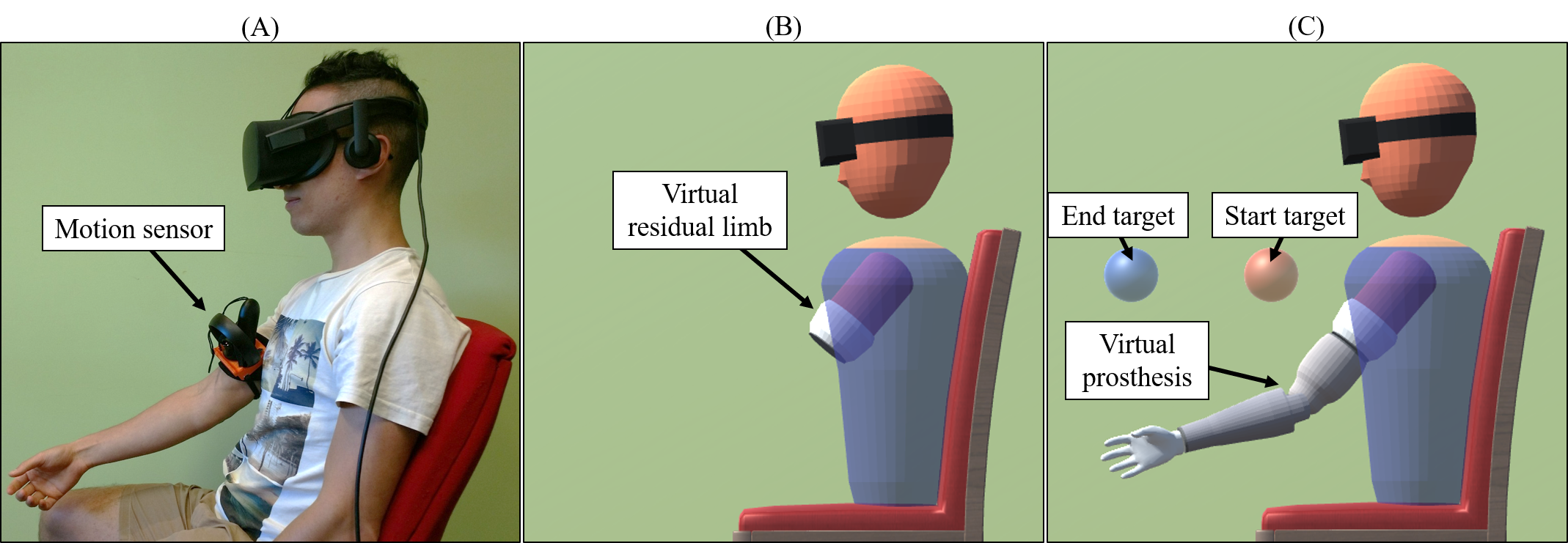}
    \vspace{-3mm}
    \caption{Virtual Reality (VR) experimental platform set-up. (A) Subject position and motor sensor placing. The motion sensor is placed on the subjects upper-arm to emulate transhumeral amputation.  (B) Virtually amputated subject in VR environment. The virtual residual limb tracks the subject's upper-arm as per the motion sensor. (C) Subject's avatar with virtual prosthesis. The lower-arm is replaced by a prosthetic device. The task requires subjects to reach between the start (red) and end (blue) targets.}
    \label{fig:setup}
    \vspace{-5mm}
\end{figure*}

The experiment required subjects to perform a center-out forward reaching task between two static targets in a Virtual Reality (VR) environment while using a virtual synergistic prosthetic elbow. The virtual targets were 23 cm apart and were placed at the subject's chest level with the start target being placed 20 cm from the target's chest. The position of the targets is shown in Figure \ref{fig:setup}.(C).
One reach motion between the two targets was considered as an iteration of the task.
Each reaching motion had to be completed within 3 seconds. Reaching attempts that were not completed within this time were discarded and had to be repeated. The task was deemed as completed if the virtual prosthetic hand stopped within 5cm of the target.
The task was performed from a seating position with trunk motion restricted, which was ensured through a ``seat belt''; shoulder and upper arm motion was unrestricted.
Transhumeral amputation was simulated by tracking only the absolute motion of the subject's dominant upper arm, coupled to the trunk and shoulder motion, which determined the motion of the virtual residual limb. Motion tracking sensor placement is shown in Figure \ref{fig:setup}.(A) while the subject's virtual avatar with residual limb is shown in Figure \ref{fig:setup}.(B).

The scalar linear kinematic synergy related to shoulder flexion/extension chosen for the HPI parameterization was given by equation (\ref{eq:linearKinematicSynergy}), with $\theta$ being the parameter to be identified. The synergy range was set as $\theta\in[0.8,2.4]$. These limits were chosen to constrain the adaptation algorithm in case of undesirable behavior. The lower limit allowed for sufficient elbow extension for the subjects to be able to perform the task while the upper limit ensured a ``safe'' speed. An initial synergy $\theta_{o} = 1.0$ was chosen for the task.

Performance of the task for a given iteration was determined by the objective function presented in equation (\ref{eq:experimentalCost}), which is related to reaching accuracy and the time to complete the reaching motion. This measure of performance was presented to the subject as a ``score'' at the end of each iteration of the task.
This was done in order to control the variable that relates to the difference between an individual's internal measure of performance and the designed function.  The first period of the dither, equivalent to 8 iterations of the task, was used to initialize the different filters and components of the algorithm; therefore, no synergy update was performed during this period. The same algorithm tuning as in the simulation implementation was used.

%
% Set-up
%
\subsection{Experiment set-up}
The experiment was performed in virtual reality using an Oculus Rift head mounted display (HMD), shown in Figure \ref{fig:setup}.A, which allows for full head tracking. Motion of the subject's dominant upper arm was tracked using an Oculus Controller sensor in order to emulate above elbow amputation as shown in Figure \ref{fig:setup}.A. The sensor allows for the absolute position and orientation of the upper limb of the human subject to be tracked in 3D space, representing the residual limb in the transhumeral amputee case that is being simulated. This means that the subject's trunk movement would also be captured in the VR simulation.  Subjects were presented with a First Person View avatar with a residual limb, which tracked the absolute motion of the subject's upper arm, and a virtual prosthetic elbow and hand as shown in Figure \ref{fig:setup}.C. A video explanation of the experimental set-up can be found in https://youtu.be/Thtjyw9lUVo.

The VR experiment platform was developed on an Oculus Rift headset with the application developed in Unity3D. The experiment was run on an Intel Core i7-7700HQ processor at 3.8GHz, with 16GB RAM, and an NVIDIA GeForce GTX 1070 video card with 8GB GDDR5. The Oculus Rift set-up included 3 base stations, two placed in the front corners of the room and one in a back corner, and one Oculus Touch controller for upper limb motion tracking. Data gathering, and VR update were performed at 90Hz.

%
% Set-up
%
\subsection{Data gathering}
Consider the following frames of reference: 
\begin{enumerate}
    \item Virtual world frame of reference. This frame of reference was chosen such that for the task the z-axis represents the forward direction, the y-axis the upwards direction, and the x-axis the lateral direction. The origin was placed on the floor under the position where the subjects were sitting to perform the task.
    \item Sensor frame of reference. This frame of reference is determined by the sensor's yaw, pitch, and roll
\end{enumerate}

A summary of the data gathered for the experiment is presented in Table \ref{ta:dataSummary}. Upper arm orientation was measured in the sensor frame of reference. The sensor was placed such that upper arm flexion/extension was represented by sensor pitch, abduction/adduction by yaw, and humeral rotation by roll. Upper arm angular velocity was measured in the sensor frame of reference. Upper arm angular velocity was represented with the same reference as upper arm orientation. The virtual hand absolute position in the virtual space was measured in the virtual world frame of reference. Finally, the absolute angular position of the virtual elbow was measured with full elbow extension as the origin.

\begin{table}[ht]
\vspace{-3mm}
    \caption{Summary of data gathered from experiments.}
    \vspace{-2mm}
    \centering
    \begin{tabular}{|c||c|c|c|c|}
        \hline
        Data & Frame of reference & x/yaw & y/pitch & z/roll  \\
        \hline
        \hline
        Upper arm & Sensor & abduction & flexion & rotation \\
        orientation & & & & \\
        \hline
        Upper arm & & & & \\
        angular & Sensor & abduction & flexion & rotation \\
        velocity & & & & \\
        \hline
        Virtual hand & Virtual world & lateral & upwards & forward \\
        position & & & & \\
        \hline
        Virtual elbow & Sensor & flexion & N/A & N/A \\
        orientation & & & & \\
        \hline
    \end{tabular}
    \label{ta:dataSummary}
    \vspace{-2mm}
\end{table}

This data was used to determine the subject's performance according to the objective function in equation (\ref{eq:experimentalCost}), and to determine the virtual prosthetic elbow motion as a function of the subject's shoulder flexion/extension.

%
% Protocol
%
\subsection{Experimental protocol}
The experiment was performed on 12 able-bodied subjects, 7 female and 5 male. The mean age and range were 30 and 19-68 respectively. Subjects received instructions on how to perform the task and performed the motion with their dominant arm before going into VR. First, the position of the sensor was calibrated, and the virtual arm was fitted to resemble their actual limb in size as close as possible as with fitting of a prosthetic device. Subjects received instructions about the task and scoring system in VR and were allowed 5 practice iterations with no time limit. Subjects had to perform the task for 150 iterations, receiving 1 minute rest every 25 iterations to minimize upper arm fatigue, and received 5 minutes rest outside VR after the $75^{th}$ iteration in order to reduce VR induced fatigue. Subjects' experience with VR environments ranged from no previous experience to casual users for gaming. No noticeable difference in performance of the task was seen among subjects with different exposure to VR. The procedure was approved by the University of Melbourne Human Research Ethics Committee, project number 1750711.1.

\subsection{Experimental results}
\begin{figure}[ht]
    \vspace{-3mm}
    \centering
    \includegraphics[trim={20pt 0pt 50pt 0pt}, width=0.4\textwidth, clip]{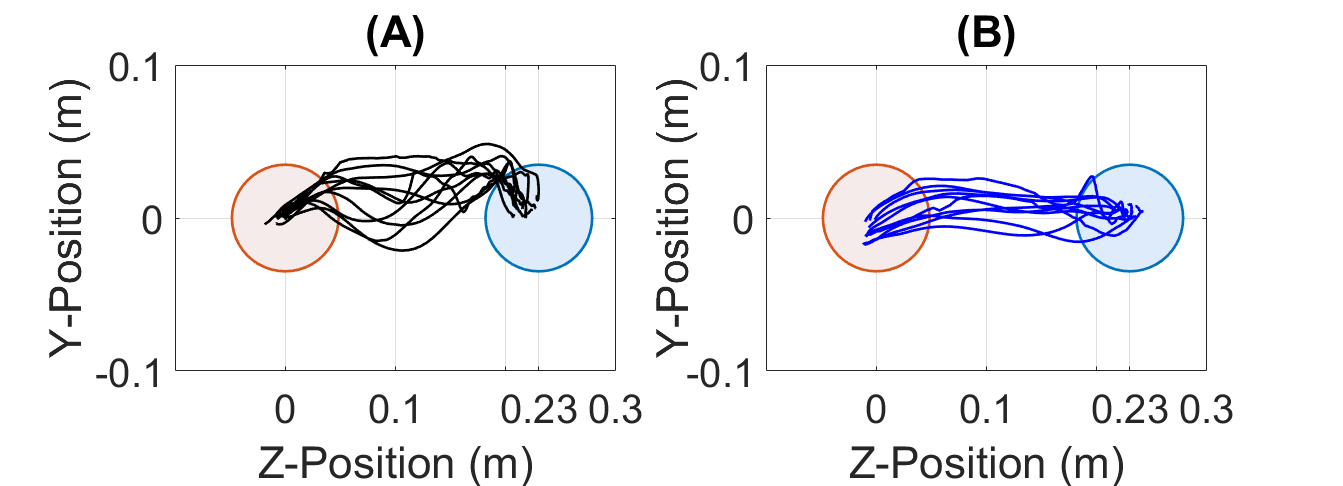}
    \vspace{-4mm}
    \caption{Representative subject's hand path in the sagittal plane. Start and end targets shown in red and blue respectively. The x-axis represents the forward direction while the y-axis the elevation. (A) The hand path for the first 10 iterations of the task. (B) The hand path for the last 10 iterations of the task.}
    \label{fig:sagittalMotion}
    \vspace{-2mm}
\end{figure}

Figure \ref{fig:sagittalMotion} shows a representative subject's hand path on the sagittal plane for the first and last 10 iterations of the forward reaching task. In the first 10 iterations of the task the synergy remained constant in order to initialize the filter, and gradient and curvature observer in the algorithm. As such, Fig. \ref{fig:sagittalMotion}.(A) shows the hand path for the case when the synergy is set at the initial value of $\theta_{o} = 1$. Fig. \ref{fig:sagittalMotion}.(B) shows the hand path of the subject when steady-state was already reached at a personalized synergy value of $\theta_{ss} = 1.5$. 

There are two main differences that can be observed in Figure \ref{fig:sagittalMotion}: the accuracy at the end point and the variance of the hand path. Given that end-point accuracy was used as part of the measure of performance (\ref{eq:experimentalCost}), it is important to highlight how the synergy can affect end-point accuracy and how it led to the observed difference. Intuitively, a value of $\theta < 1$ means the elbow extends less than the shoulder and vice versa. A larger $\theta$ also means a larger change in elbow extension for a change in the shoulder flexion, which makes it easier to reach further (thus to reach the target) but at the same time increases the difficulty to control the position of the hand. Therefore, as $\theta$ increases, the overall performance increases until the subjects are able to achieve their best accuracy, which will be limited by their individual motor behavior. The difference between Fig. \ref{fig:sagittalMotion}.(A) and (B) with respect to the variability of the hand path may be attributed to human motor learning \cite{Darling1987}.

\begin{figure}[ht]
    \centering
    \vspace{-3mm}
    \includegraphics[trim={50pt 0pt 50pt 0pt}, width=0.4\textwidth, clip]{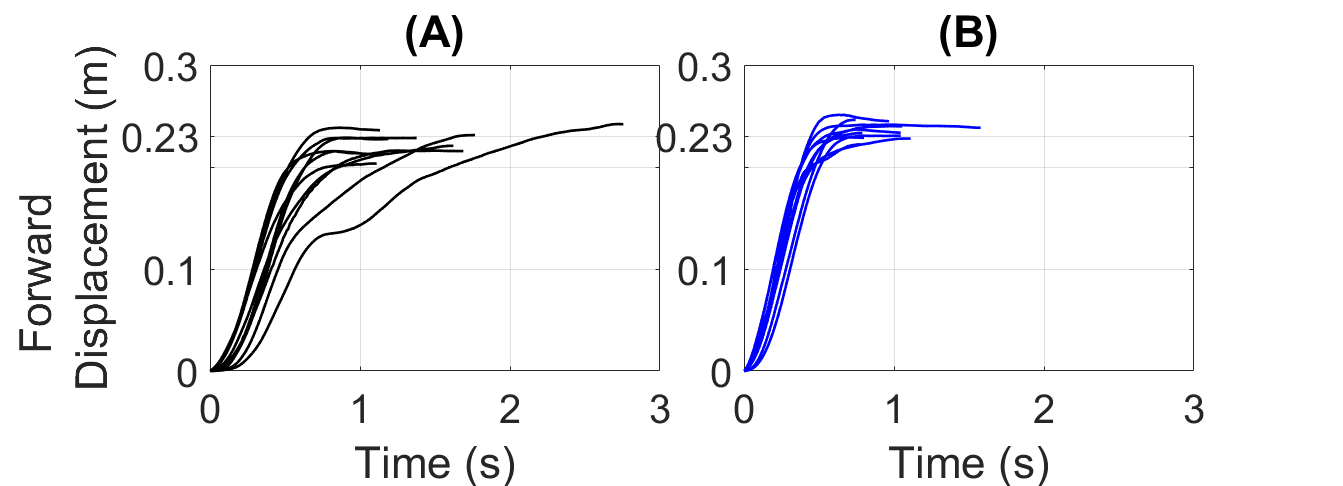}
    \vspace{-4mm}
    \caption{Representative subject's hand forward displacement across time. (A) The hand forward displacement for the first 10 iterations of the task. (B) The hand forward displacement for the last 10 iterations of the task.}
    \label{fig:forwardMotion}
    \vspace{-2mm}
\end{figure}

Figure \ref{fig:forwardMotion} shows a representative subject's hand forward displacement over time for the first and last 10 iterations of the forward reaching task. Similarly to Fig. \ref{fig:sagittalMotion}, Fig. \ref{fig:forwardMotion} shows how the synergy affects one of the components of the measure of performance (\ref{eq:experimentalCost}), namely the task completion time. Again, Fig. \ref{fig:forwardMotion}.(A) shows the case of the initial synergy $\theta_{o} = 1$, and Fig. \ref{fig:forwardMotion}.(B) the case of the personalized synergy $\theta_{ss} = 1.5$. It is important to highlight in this figure that the rise time and task completion time are both reduced after the synergy reaches steady-state ($\theta \rightarrow \theta_{ss}$). 

Intuitively, a larger synergy value allows the elbow to travel faster, and thus the hand, for a given shoulder flexion, in turn reducing the task time. However, given that the synergy also affects hand displacement, a value that is too large will lead to over-extension and a decay in accuracy. Thus, there is a trade-off between speed and accuracy, which is determined by the objective function (equation (\ref{eq:experimentalCost})), and determines the optimal personalized synergy. Given that different people have different motor behavior and capabilities, the best synergy value will differ across individuals.

\begin{figure*}[ht!]
    \centering
    %\vspace{-0mm}
    %\includegraphics[trim={1.6in 0.3in 1.7in 0.3in}, width=6.0in, clip]{Figures/thetaResults.eps}
    \includegraphics[width=7.0in]{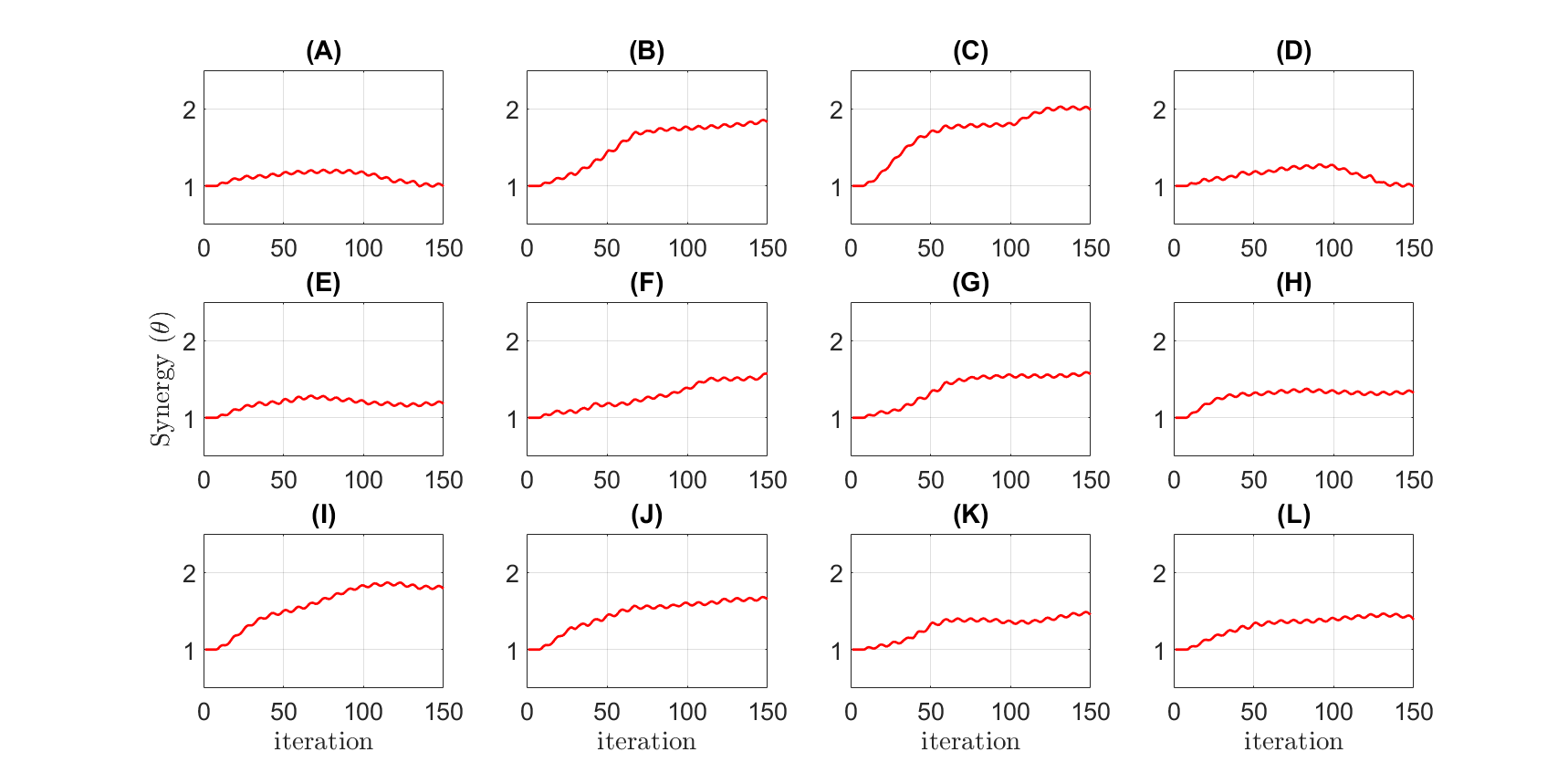}
    \vspace{-5mm}
    \caption{Synergy value across iterations of the task for all subjects in the experiment. The personalization algorithm was capable of uniformly adapting and identifying the synergy to allow all subjects to improve and reach a steady-state performance within the 150 iterations regardless of their particular motor preference and adaptation characteristics.}
    \label{fig:thetaResults}
    \vspace{-5mm}
\end{figure*}

\begin{figure*}[ht!]
    \centering
    %\vspace{-5mm}
    %\includegraphics[trim={1.6in 0.3in 1.7in 0.3in}, width=6.0in, clip]{Figures/costResults.eps}
    \includegraphics[width=7.0in]{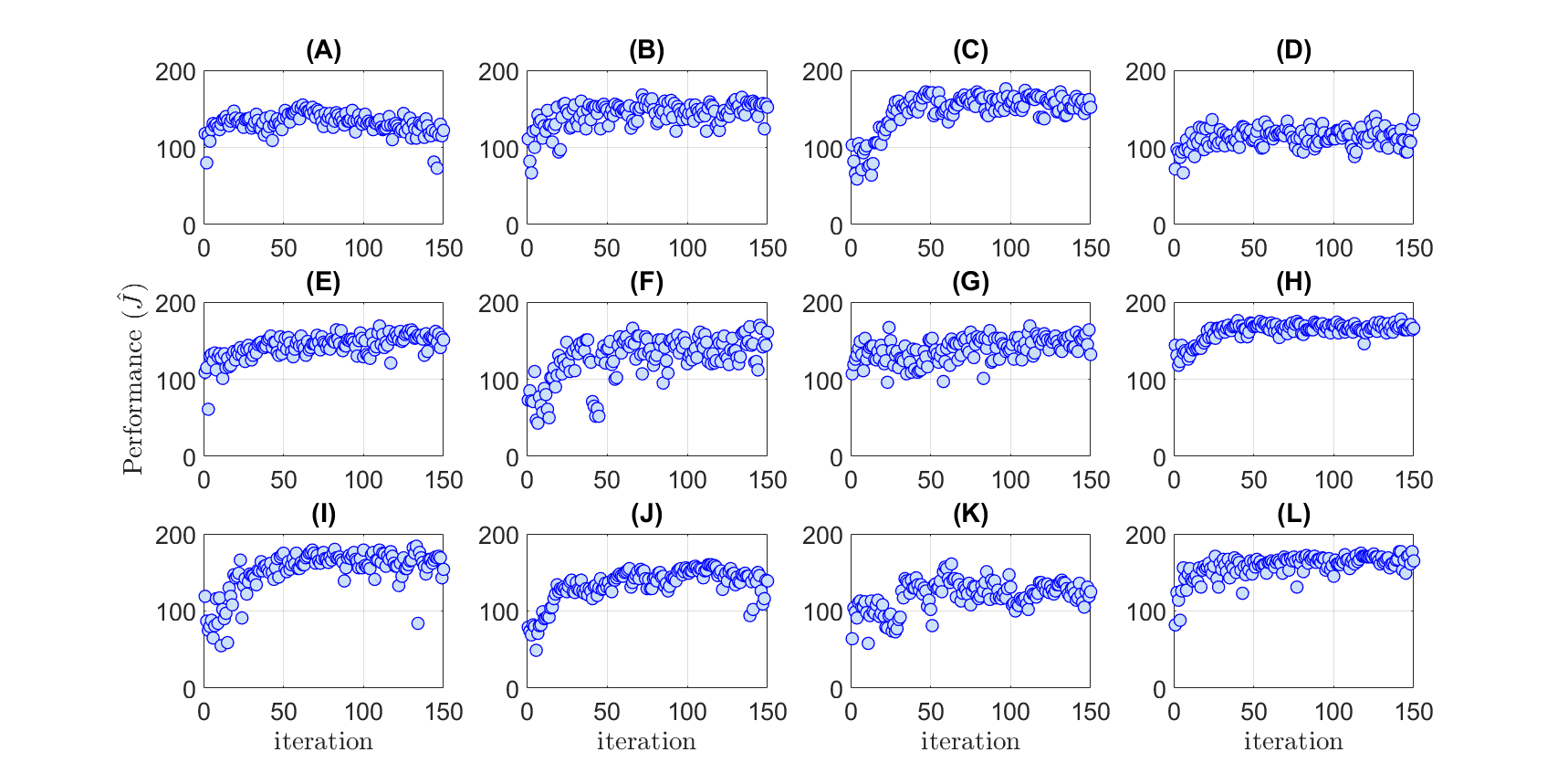}
    \vspace{-5mm}
    \caption{Performance across iterations for all subjects in the experiment, where one reach motion between the two targets was considered as an iteration of the task. A wide variety of capabilities can be observed among the different subjects, such as multiple different performance steady-states and levels of performance variability.}
    \label{fig:costResults}
    \vspace{-6mm}
\end{figure*}

The change in synergy across the 150 iterations of the task are presented in Figure \ref{fig:thetaResults}. This shows how the personalization algorithm adapted the synergy for each individual until reaching a steady-state, which is considered as the personalized synergy. Each subject's performance across the 150 iterations of the task is presented in Figure \ref{fig:costResults}. Each data point in Fig. \ref{fig:costResults} represents the subject's performance as defined in equation (\ref{eq:experimentalCost}) for a given reaching attempt. Together with Fig. \ref{fig:thetaResults}, Fig. \ref{fig:costResults} shows how the change in synergy led to a change in performance. In general, it can be observed that an increase in the synergy value led to an increase in performance. As previously discussed in Figs. \ref{fig:sagittalMotion} and \ref{fig:forwardMotion}, this is expected as a larger synergy value means that further elbow extension will be performed w.r.t. shoulder extension, allowing the hand to travel further and thus closer to the target; and to travel faster, thus reducing the task time.
    
% Final results

Regarding the performance of the proposed personalization algorithm itself, it can be observed in Fig. \ref{fig:costResults} that performance steady-state was reached under 50 iterations on average, which was achieved uniformly across most subjects. In other approaches, such as \cite{Garcia-Rosas2018ACC}, the performance of the algorithm depends on the motor performance map and the tuning of the algorithm to accommodate individual motor learning characteristics. However, the personalization algorithm proposed herein achieved this uniform performance regardless of the individuals' motor behavior and learning characteristics without the need to re-tune. While motor learning characteristics are difficult to discern from Fig. \ref{fig:costResults}, individual differences motor behavior can be seen in the initial and steady-state (best) performance, as well as in the relationship between the synergy Fig. \ref{fig:thetaResults} and performance Fig. \ref{fig:costResults}. The shape of the relationship between synergy and performance differs across individuals. For instance, this synergy-performance relationship can be quite flat which can cause gradient-based optimization algorithms, as in \cite{Garcia-Rosas2018ACC}, to respond slowly. A case of such relationship can be seen in Figures \ref{fig:thetaResults} and \ref{fig:costResults} in subject (G), where a large change in synergy lead to a small change in performance. However, the proposed personalization algorithm was able to adapt its update rate, through the proposed optimizer, to accommodate for this synergy-performance relationship as can be seen by the number of iterations needed to reach convergence.

A wide variety of capabilities are observed among the different subjects which is expected of human behavior. Multiple different performance steady-states with different identified best synergies can be seen across subjects. This further highlights the person to person variations, and the importance and need of personalization in human-prosthesis interfaces. Motor performance variation ($\bar{v}$) is another clear indicator of individuality, with different people having different levels of variability. In subjects with small identified $\bar{v}$, very clear performance steady-state and convergence of synergy are observed, such as subjects (H) and (L). On the other hand, in subjects with higher $\bar{v}$, the performance steady-state is not as clear and a wider drift in synergy is observed, such as with subjects (F) and (K). This shows how human motor behavior variability can affect the performance of the algorithm and motivates further improvements to algorithm to address this aspect of human behavior.

In light of the observed diversity in behavior, the personalization algorithm was capable of adapting the synergy regardless of individual motor learning characteristics to allow all subjects to improve their performance while using the same algorithm tuning across all subjects. This demonstrates the advantages of the proposed algorithm for on-line personalization of synergistic HPIs.

%
% Discussion
%
\section{Discussion}
This section discusses the applicability of the proposed personalization algorithm to other tasks and prosthetic interfaces. This algorithm is based on a general framework that can be used in a wide variety of applications through the choice of the objective function and system parametrization, which define the task and the parameters to be identified respectively. 
As such, these components will play an important role in the applicability of the personalization algorithm to practical scenarios, each presenting its own challenges. In the case of the objective function, the challenge is the formulation of a significant and practical measure of performance; in the case of the parametrization, it is in the synergy's dependency to the task. These two challenges are discussed next.

\begin{figure*}[ht!]
    \centering
    %\vspace{-0mm}
    \includegraphics[width=5.0in]{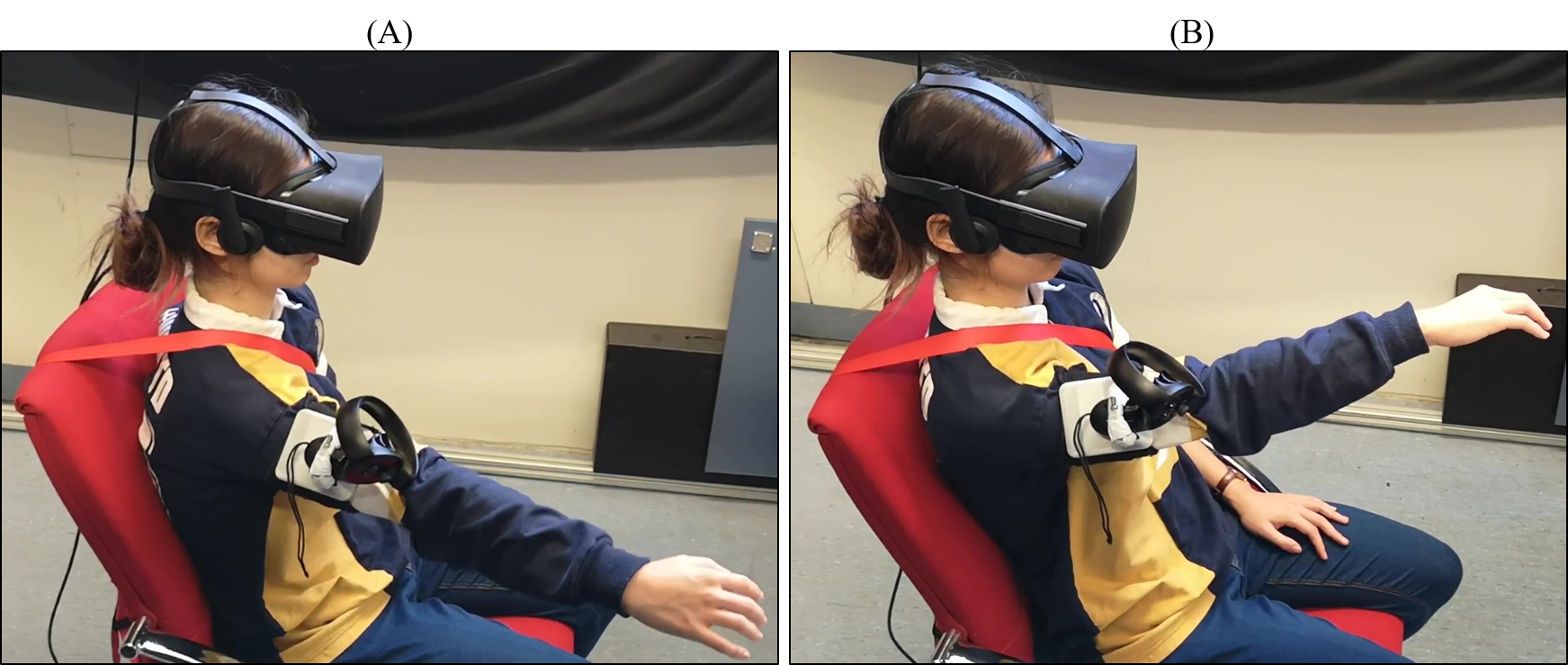}
    \vspace{-3mm}
    \caption{Subject's difference in arm position at the end of reach for a representative iteration of the first and last 10 iterations of the task. A difference in shoulder protraction can be observed. (A) Arm end position for a representative iteration within the first 10 iterations of the experiment. (B) Arm end position for a representative iteration within the last 10 iterations of the experiment.}
    \label{fig:motorBehavior}
    \vspace{-5mm}
\end{figure*}

Even though accuracy and completion time are common measures of upper limb motion performance, they rely on knowledge of the reaching task, such as object location, which would be difficult to measure. Formulation of an objective function to evaluate performance in a practical application is paramount to the success of the proposed personalization method. Given that obtaining the desired target object information from the prosthesis user can be difficult, it may be possible to use measurements of the user's upper body motion to determine the performance of the prosthesis as an assistive device instead. Thus, changing the focus of the objective function, and thus the measure of performance, to be human-behavior centered, rather than task-space centered.
For instance, upper-body motion can be used to determine the compensation motion an amputee needs in order to achieve a task with a prosthetic device \cite{Metzger2012}. Moreover, it has been observed that synergistic HPIs can reduce compensatory motion \cite{Merad2018}. In the study presented in this paper, a change in the upper-body motion of the subjects due to the personalized synergy was anecdotally observed. An example of this behavior is presented in Figure \ref{fig:motorBehavior}, where the subject's arm position at the end of the reaching task is shown for a representative reach of the first and last 10 iterations. The figure shows a difference in shoulder protraction between non-personalized and personalized synergy. As such, it is the authors' opinion that compensation motion may be a good candidate to evaluate performance of the prosthetic device and could be used to formulate an objective function for the personalization algorithm. This will be quantitatively studied in future work.

% Motor behavior changes.

In the case of the parametrization of the HPI, a significant challenge with the use of synergies is their evident dependence on the given task. Hence the difficulty in generalizing the personalized synergy from a particular point-to-point task to multiple directions, or more complex tasks. A possibility to address this problem is to consider higher-dimensional synergies or more complex synergy parametrizations in order to take advantage of the redundancies in the human arm to determine the intended direction of motion. Furthermore, other HPI modalities could be included in the parametrization in order to compliment the information obtained from the synergy with explicit user intention, such as how synergies and sEMG were combined in \cite{Alshammary2018}.
%
% Conclusion
%
\section{Conclusion}
In this paper, an algorithm for on-line personalization of kinematic synergies for Human-Prosthesis Interfaces whose tuning is independent of individual motor behavior and learning was proposed. The algorithm is able to achieve uniform convergence performance across the population.
In order to achieve this, the algorithm takes advantage of common motor preference and adaptation behavior observed experimentally in a group of individuals using a synergistic upper limb prosthetic device in a virtual reality environment. This motor behavior was formally characterized with a model structure which captures the common behavior across a population.

The application of the proposed personalization algorithm was undertaken in a virtual reality environment with able-bodied subjects by emulating amputation and prosthesis use. The algorithm was capable of adapting the synergy while subjects used the prosthetic device in a forward reaching task, allowing all subjects to improve their performance despite the demonstrated person-to-person variations. Convergence performance of the algorithm was uniform across subjects using the same parameter tuning, avoiding the need for manual tuning for each individual. 

Further work will look into quantifying the effects of synergistic prosthesis interface personalization on compensation motion and the effects of the choice of objective function for personalization on motor behavior and performance, the use of more complex synergy models that take advantage of the redundancies in the human arm, the application of the algorithm to other interface modalities, and the development of a practical implementation of the proposed personalization framework on a prototype prosthetic device with wearable motion sensors.

%% ----------------------------------------------------------------
\bibliographystyle{ieeetr}  % Use the "unsrtnat" BibTeX style for formatting the Bibliography
\bibliography{library}  % The references (bibliography) information are stored in the file named "Bibliography.bib"

\begin{thebibliography}{10}

\bibitem{Gottlieb1996}
G.~L. Gottlieb, Q.~Song, D.~A. Hong, G.~L. Almeida, and D.~Corcos,
  ``{Coordinating movement at two joints: A principle of linear covariance.},''
  {\em J. Neurophysiol.}, vol.~75, no.~4, pp.~1760--1764, 1996.

\bibitem{Cordella2016}
F.~Cordella, A.~L. Ciancio, R.~Sacchetti, A.~Davalli, A.~G. Cutti,
  E.~Guglielmelli, and L.~Zollo, ``{Literature review on needs of upper limb
  prosthesis users},'' {\em Front. Neurosci.}, vol.~10, 2016.

\bibitem{Alshammary2018}
N.~A. Alshammary, D.~A. Bennett, and M.~Goldfarb, ``{Synergistic elbow control
  for a myoelectric transhumeral prosthesis},'' {\em IEEE Trans. Neural Syst.
  Rehabil. Eng.}, vol.~26, no.~2, pp.~468--476, 2018.

\bibitem{Legrand2018}
M.~Legrand, M.~Merad, E.~de~Montalivet, A.~Roby-Brami, and N.~Jarrass{\'{e}},
  ``{Movement-based control for upper-limb prosthetics: Is the regression
  technique the key to a robust and accurate control?},'' {\em Front.
  Neurorobot.}, vol.~12, 2018.

\bibitem{Kaliki2013}
R.~R. Kaliki, R.~Davoodi, and G.~E. Loeb, ``{Evaluation of a noninvasive
  command scheme for upper-limb prostheses in a virtual reality reach and grasp
  task},'' {\em IEEE Trans. Biomed. Eng.}, vol.~60, no.~3, pp.~792--802, 2013.

\bibitem{Blana2016}
D.~Blana, T.~Kyriacou, J.~M. Lambrecht, and E.~K. Chadwick, ``{Feasibility of
  using combined EMG and kinematic signals for prosthesis control: A simulation
  study using a virtual reality environment},'' {\em J. Electromyogr.
  Kinesiol.}, vol.~29, pp.~21--27, 2016.

\bibitem{Merad2016a}
M.~Merad, E.~de~Montalivet, A.~Roby-Brami, and N.~Jarrasse, ``{Intuitive
  prosthetic control using upper limb inter-joint coordinations and IMU-based
  shoulder angles measurement: A pilot study},'' in {\em Int. Conf. Intell.
  Robot. Syst.}, pp.~5677--5682, 2016.

\bibitem{Akhtar2017}
A.~Akhtar, N.~Aghasadeghi, L.~Hargrove, and T.~Bretl, ``{Estimation of distal
  arm joint angles from EMG and shoulder orientation for transhumeral
  prostheses},'' {\em J. Electromyogr. Kinesiol.}, vol.~35, pp.~86--94, 2017.

\bibitem{Merad2018}
M.~Merad, {\'{E}}.~de~Montalivet, A.~Touillet, N.~Martinet, A.~Roby-Brami, and
  N.~Jarrass{\'{e}}, ``{Can we achieve intuitive prosthetic elbow control based
  on healthy upper limb motor strategies?},'' {\em Front. Neurorobot.},
  vol.~12, 2018.

\bibitem{Biddiss2007a}
E.~A. Biddiss and T.~T. Chau, ``{Upper limb prosthesis use and abandonment: A
  survey of the last 25 years},'' {\em Prosthet. Orthot. Int.}, vol.~31, no.~3,
  pp.~236--257, 2007.

\bibitem{Fougner2012}
A.~Fougner, O.~Stavdahl, P.~J. Kyberd, Y.~G. Losier, and P.~A. Parker,
  ``{Control of upper limb prostheses: Terminology and proportional myoelectric
  control - A review},'' {\em IEEE Trans. Neural Syst. Rehabil. Eng.}, vol.~20,
  no.~5, pp.~663--677, 2012.

\bibitem{Garcia-Rosas2018EMBC}
R.~Garcia-Rosas, D.~Oetomo, C.~Manzie, Y.~Tan, and P.~Choong, ``{On the
  relationship between human motor control performance and kinematic synergies
  in upper limb prosthetics},'' in {\em Eng. Med. Biol. Conf.}, pp.~3194 --
  3197, 2018.

\bibitem{Garcia-Rosas2018ACC}
R.~Garcia-Rosas, Y.~Tan, D.~Oetomo, and C.~Manzie, ``{On-line synergy
  identification for personalized active arm prosthesis: A feasibility
  study},'' in {\em Am. Control Conf.}, pp.~514--519, 2018.

\bibitem{Kumar2017}
S.~Kumar, A.~Mohammadi, N.~Gans, and R.~D. Gregg, ``{Automatic tuning of
  virtual constraint-based control algorithms for powered knee-ankle
  prostheses},'' in {\em Conf. Control Technol. Appl.}, pp.~812--818, 2017.

\bibitem{Moase2011a}
W.~H. Moase and C.~Manzie, ``{Fast extremum-seeking on Hammerstein plants},''
  {\em IFAC Proc. Vol.}, vol.~44, no.~1, pp.~108--113, 2011.

\bibitem{Moase2009a}
W.~H. Moase, C.~Manzie, and M.~J. Brear, ``{Newton-like extremum-seeking part
  I: Theory},'' in {\em IEEE Conf. Decis. Control}, pp.~3839--3844, 2009.

\bibitem{Bockemuhl2010}
T.~Bockem{\"{u}}hl, N.~F. Troje, and V.~D{\"{u}}rr, ``{Inter-joint coupling and
  joint angle synergies of human catching movements},'' {\em Hum. Mov. Sci.},
  vol.~29, no.~1, pp.~73--93, 2010.

\bibitem{Riley2002}
M.~A. Riley and M.~T. Turvey, ``{Variability and determinism in motor
  behavior},'' {\em J. Mot. Behav.}, vol.~34, no.~2, pp.~99--125, 2002.

\bibitem{Cheng2006}
S.~Cheng and P.~N. Sabes, ``{Modeling sensorimotor learning with linear
  dynamical systems},'' {\em Neural Comput.}, vol.~18, no.~4, pp.~760--793,
  2006.

\bibitem{Wei2010}
K.~Wei and K.~Kording, ``{Uncertainty of feedback and state estimation
  determines the speed of motor adaptation},'' {\em Front. Comput. Neurosci.},
  2010.

\bibitem{He2016}
K.~He, Y.~Liang, F.~Abdollahi, M.~{Fisher Bittmann}, K.~Kording, and K.~Wei,
  ``{The statistical determinants of the speed of motor learning},'' {\em PLOS
  Comput. Biol.}, vol.~12, no.~9, 2016.

\bibitem{Zhou2012}
S.-H. Zhou, D.~Oetomo, Y.~Tan, E.~Burdet, and I.~Mareels, ``{Modeling
  individual human motor behavior through model reference iterative learning
  control},'' {\em IEEE Trans. Biomed. Eng.}, vol.~59, no.~7, pp.~1892--901,
  2012.

\bibitem{Johnson2014a}
R.~E. Johnson, K.~P. Kording, L.~J. Hargrove, and J.~W. Sensinger, ``{Does EMG
  control lead to distinct motor adaptation?},'' {\em Front. Neurosci.},
  vol.~8, 2014.

\bibitem{Shehata2017}
A.~W. Shehata, E.~J. Scheme, and J.~W. Sensinger, ``{The effect of myoelectric
  prosthesis control strategies and feedback level on adaptation rate for a
  target acquisition task},'' in {\em 2017 Int. Conf. Rehabil. Robot.},
  pp.~200--204, 2017.

\bibitem{Landau2011}
I.~D. Landau, R.~Lozano, M.~M'Saad, and A.~Karimi, {\em {Adaptive control}}.
\newblock Springer, 2011.

\bibitem{Nesic2010a}
D.~Nesic, Y.~Tan, W.~H. Moase, and C.~Manzie, ``{A unifying approach to
  extremum seeking: adaptive schemes based on estimation of derivatives},''
  {\em IEEE Conf. Decis. Control}, pp.~4625--4630, 2010.

\bibitem{Khalil2002}
H.~K. Khalil and J.~Grizzle, {\em {Nonlinear systems}}.
\newblock Pearson, 3rd~ed., 2002.

\bibitem{Ariyur2003}
K.~B. Ariyur and M.~Krsti{\'{c}}, {\em {Real-time optimization by
  extremum-seeking control}}.
\newblock John Wiley {\&} Sons, 2003.

\bibitem{Cardis2017}
M.~Cardis, M.~Casadio, and R.~Ranganathan, ``{High variability impairs motor
  learning regardless of whether it affects task performance},'' {\em J.
  Neurophysiol.}, 2017.

\bibitem{Darling1987}
W.~G. Darling and J.~D. Cooke, ``{Changes in the variability of movement
  trajectories with practice},'' {\em J. Mot. Behav.}, vol.~19, no.~3,
  pp.~291--309, 1987.

\bibitem{Metzger2012}
A.~J. Metzger, A.~W. Dromerick, R.~J. Holley, and P.~S. Lum,
  ``{Characterization of compensatory trunk movements during prosthetic upper
  limb reaching tasks},'' {\em Arch. Phys. Med. Rehabil.}, vol.~93, no.~11,
  pp.~2029--2034, 2012.

\end{thebibliography}

\begin{IEEEbiography}[{\includegraphics[width=1in,height =1.25in,clip,keepaspectratio]{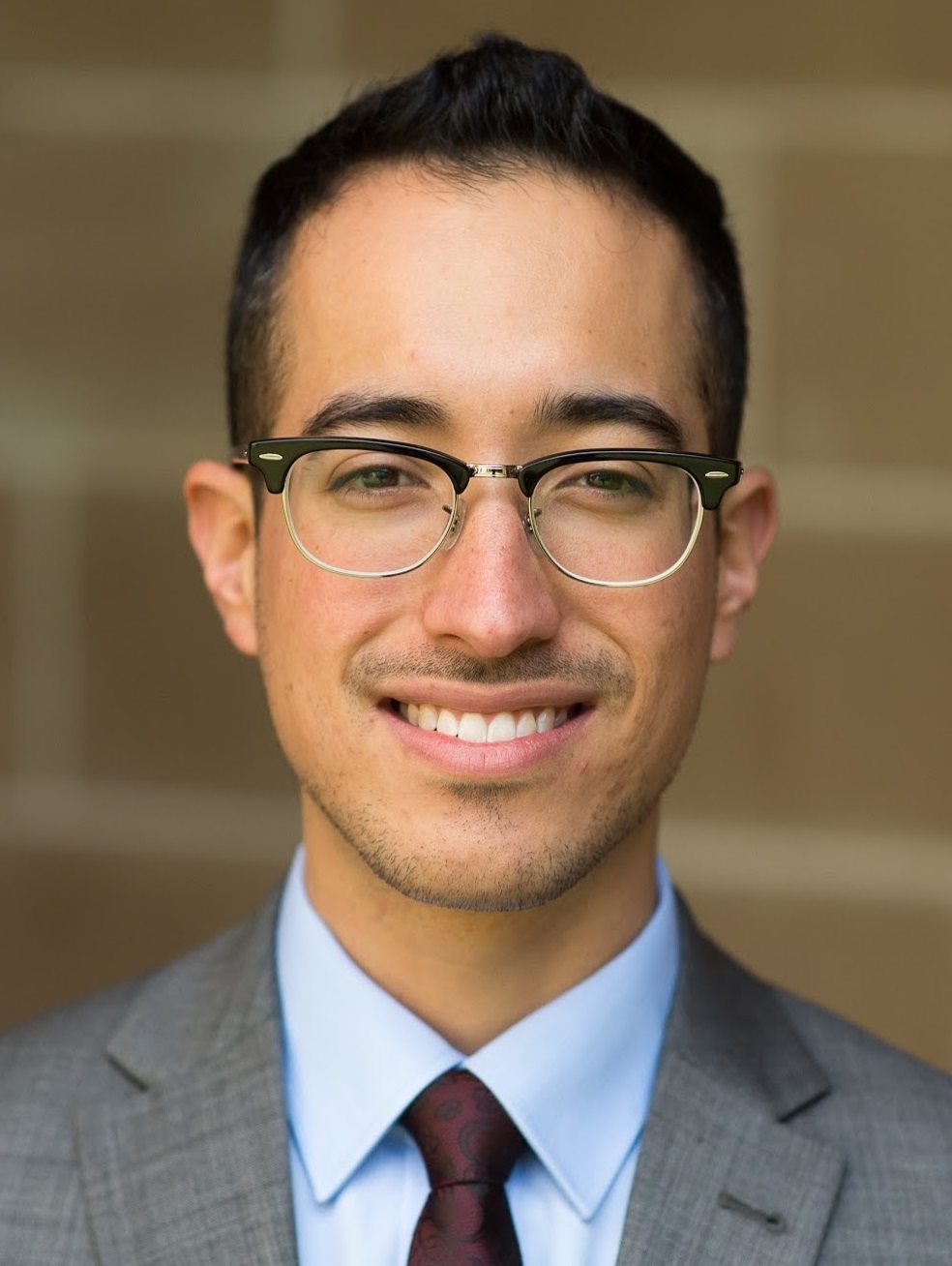}}]{Ricardo Garcia-Rosas} is a PhD candidate in the Melbourne Human-Robotics Lab at the University of Melbourne, Australia. He obtained his B.Eng. in Mechatronics in 2010 from Tecnologico de Monterrey, Mexico; and his M.Eng. in Mechatronics (with Distinction) in 2016 from the University of Melbourne, Australia. His research aims to autonomously personalize the behaviour of smart upper limb prostheses to that of particular individuals in order to enable complex and coordinated human-robot arm motion. \end{IEEEbiography}

\begin{IEEEbiography}[{\includegraphics[width=1in,height =1.25in,clip,keepaspectratio]{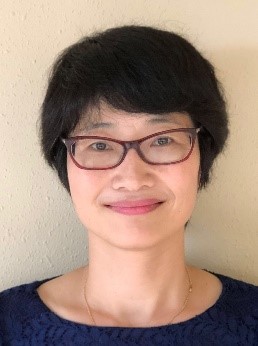}}]{Ying Tan} is an Associate Professor and Reader in the Department of Electrical and Electronic Engineering (DEEE) at the University of Melbourne, Australia. She received her Bachelor’s degree from Tianjin University, China, in 1995, and her PhD from the Department of Electrical and Computer Engineering National University of Singapore in 2002. She joined McMaster University in 2002 as a postdoctoral fellow in the Department of Chemical Engineering. Since 2004, she has been with the University of Melbourne.  She was awarded an Australian Postdoctoral Fellow (2006-2008) and a Future Fellow (2009-2013) by the Australian Research Council.  Her research interests are in intelligent systems, nonlinear control systems, real time optimisation, sampled-data distributed parameter systems and formation control.  \end{IEEEbiography}

\vspace{-5mm}
\begin{IEEEbiography}[{\includegraphics[width=1in,height =1.25in,clip,keepaspectratio]{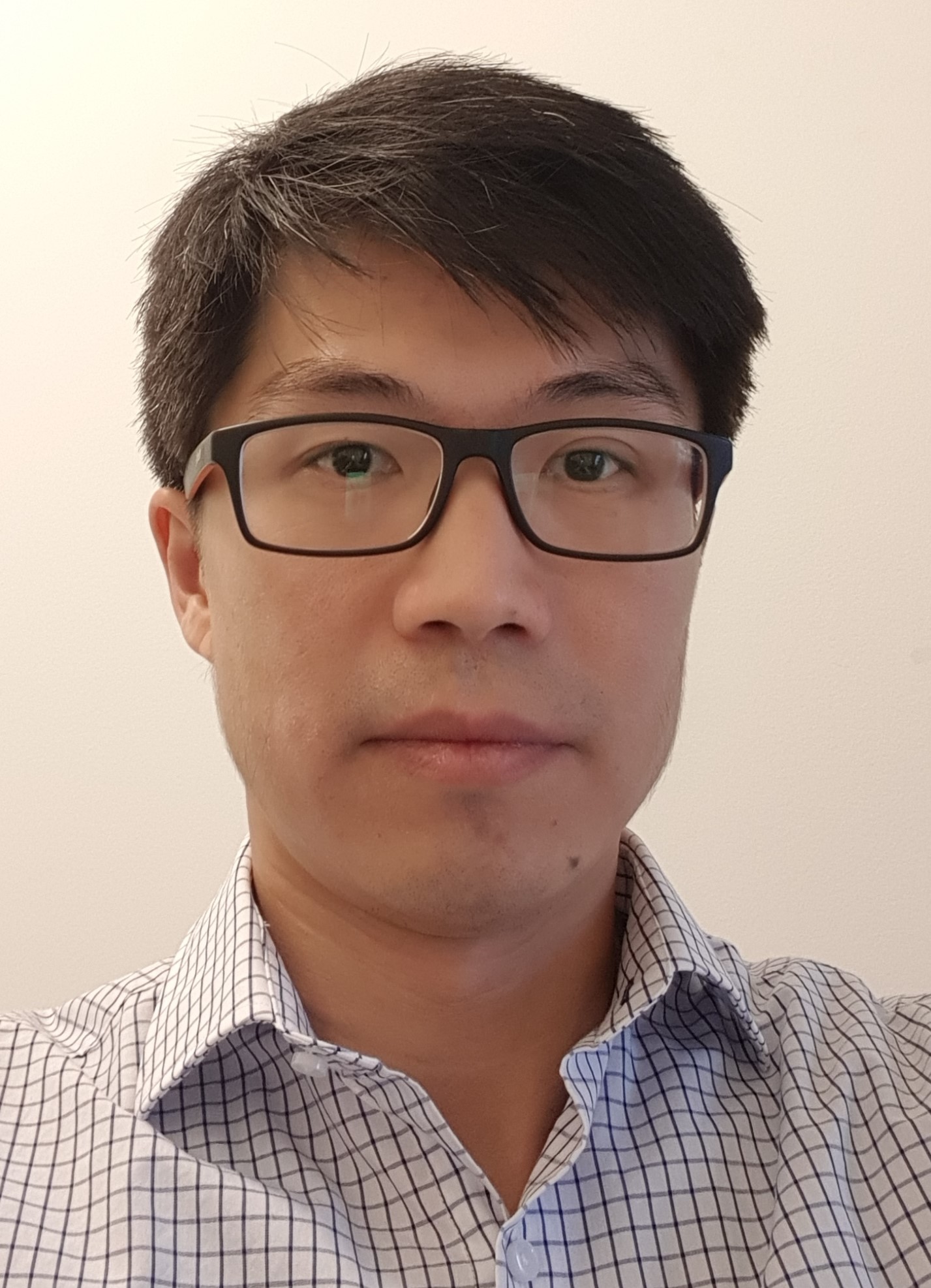}}]{Denny Oetomo} graduated BE (Hons 1, 1997) from the Australian National University and PhD (Mechanical Eng, Robotics, 2004) from the National University of Singapore. After postdoctoral fellowships in Monash University and INRIA Sophia-Antipolis, he joined the University of Melbourne in 2008 where he is currently an Associate Professor. His research interests are in the areas of the manipulation of robot dynamics, through design and manipulation strategies, with the more recent emphasis on (physical) human assistive robots, applied in areas of clinical rehabilitation, prosthetics and industrial robotics.  \end{IEEEbiography}

\vspace{-5mm}
\begin{IEEEbiography}[{\includegraphics[width=1in,height =1.25in,clip,keepaspectratio]{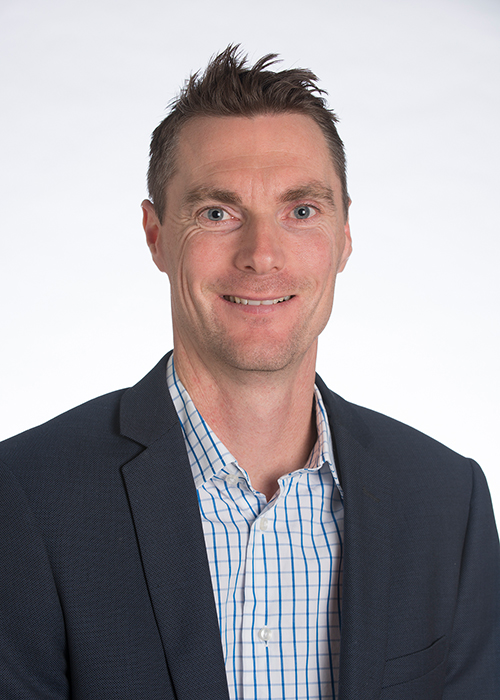}}]{Chris Manzie} is Head of Department of Electrical and Electronic Engineering at the University of Melbourne, and also the Director of the Melbourne Information, Decision and Autonomous Systems (MIDAS) Laboratory, which includes academics from multiple faculties including Engineering, Science and Law. Over the period 2003-2016, he was an academic in the Department of Mechanical Engineering, with responsibilities including Assistant Dean with the portfolio of Research Training (2011-2017), and Mechatronics Program Director (2009-2016). Professor Manzie has had visiting positions with the University of California, San Diego and IFP Energies Nouvelles, Rueil Malmaison.  His research interests are in model-based and model-free control and optimisation, with applications in a range of areas including systems related to autonomous systems, energy, transportation and mechatronics. He is currently an Associate Editor for IEEE Transactions on Control Systems Technology and Elsevier Mechatronics and is a past Associate Editor for Control Engineering Practice and IEEE/ASME Transactions on Mechatronics. \end{IEEEbiography}

\vspace{-5mm}
\begin{IEEEbiography}[{\includegraphics[width=1in,height =1.25in,clip,keepaspectratio]{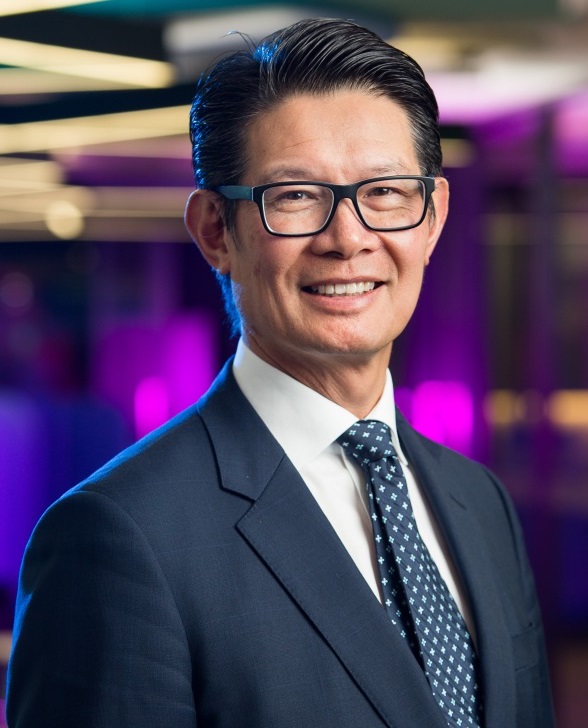}}]{Peter Choong} is the Sir Hugh Devine Professor of Surgery, and Head of Department of Surgery University of Melbourne at St. Vincent’s hospital Melbourne.  Peter was the inaugural Professor of Orthopaedics (Adult) in Melbourne and is also the Director of Orthopaedics at St. Vincent’s Hospital Melbourne. He is the Chair of the Bone and Soft Tissue Sarcoma Service at the Peter MacCallum Cancer Centre and is the recent past President of the Australian Orthopaedic Association.

Peter’s research which has been supported by recurrent grant funding of almost \$25million (NHMRC, ARC, RACS, AOA, Cancer Australia, CRC, MRFF, NGO) and focuses on 3 areas, that underpin his clinical programmes, namely improving outcomes of arthritis surgery, studying the treatment of bone tumours and advanced limb reconstruction. Prof Choong is widely published with almost 400 peer-reviewed articles (H score 64) in these areas. In 2014, he was elected a Fellow of the Australian Academy of Health and Medical Science and in 2019, he was awarded an National Health and Medical Research Council Practitioner Fellowship for which he was also awarded a Research Excellence award for being the highest ranked applicant. Peter is the Lead Chief Investigator on an NHMRC Centre for Research Excellence(CRE) into Total Joint Replacement Surgery Research and is the Principle Clinical Lead in the Innovative Manufacturing Cooperative Research Centre (CRC).
 \end{IEEEbiography}

\end{document}